\pdfoutput=1

\documentclass[11pt]{article}

\usepackage[final]{acl}
\usepackage{times}
\usepackage{latexsym}
\usepackage[T1]{fontenc}
\usepackage[utf8]{inputenc}
\usepackage{microtype}
\usepackage{inconsolata}
\usepackage{graphicx}

\usepackage{amsmath,amsfonts,bm}

\def\eqref#1{equation~\ref{#1}}

\def\1{\bm{1}}

\DeclareMathAlphabet{\mathsfit}{\encodingdefault}{\sfdefault}{m}{sl}
\SetMathAlphabet{\mathsfit}{bold}{\encodingdefault}{\sfdefault}{bx}{n}

\usepackage{amsmath}
\usepackage{bbm}
\usepackage{amssymb}
\usepackage{mathtools}
\usepackage{amsthm}

\usepackage{algorithm}
\usepackage{algorithmic}
\usepackage{setspace}
\usepackage{comment}
\usepackage{booktabs}
\usepackage{hyperref}
\usepackage{icomma}
\usepackage{url}

\usepackage[nameinlink]{cleveref}
\usepackage{longtable}
\usepackage{xcolor}
\usepackage{colortbl}
\usepackage{multicol}
\usepackage{multirow}
\usepackage{makecell}
\usepackage{amsmath, amssymb}
\usepackage{mathtools}
\usepackage{enumitem}
\usepackage{subcaption}
\usepackage{float}
\usepackage{graphicx}
\usepackage{caption} 
\usepackage{upgreek}
\usepackage{seqsplit}
\usepackage{color,soul}
\usepackage{arydshln}
\usepackage{xspace}
\usepackage{adjustbox}
\usepackage{enumitem}
\usepackage{booktabs} 
\usepackage[most]{tcolorbox}

\usepackage{pifont}
\newcommand{\cmark}{\textcolor{green!60!black}{\ding{51}}}
\newcommand{\xmark}{\textcolor{red}{\ding{55}}}

\definecolor{Red}{rgb}{0.6,0,0}
\definecolor{Blue}{rgb}{0,0,0.8}
\definecolor{Green}{rgb}{0,0.4,0.7}
\definecolor{mountainmeadow}{rgb}{0.19, 0.73, 0.56}
\definecolor{crimson}{rgb}{0.86, 0.08, 0.24}
\definecolor{darkblue}{rgb}{0.0, 0.0, 0.60}
\definecolor{umber}{RGB}{117,79,41}

\definecolor{gg}{HTML}{E0FEFE}
\definecolor{gray}{RGB}{236, 236, 236}

\hypersetup{
    colorlinks = true,
    citecolor = darkblue,
    linkcolor = darkblue,
    linktoc = all,
    urlcolor = darkblue,
}

\newcommand{\highlight}[1]{{\color{darkblue}{#1}}}
\newcommand{\ours}{\textsc{HiViG}\xspace}

\newcommand{\myparagraph}[1]{\vspace{0.25em}\noindent \textbf{#1}\hspace{0.5em}}

\title{A History-Aware Visually Grounded Critic for Computer Use Agents}

\author{
    Jaewoo Lee$^{1}$ \;
    Zaid Khan$^{1}$ \;
    Archiki Prasad$^{1}$ \;
    Justin Chih-Yao Chen$^{1}$ \\
    \textbf{Supriyo Chakraborty}$^{2}$ \;
    \textbf{Kartik Balasubramaniam}$^{2}$ \;
    \textbf{Sambit Sahu}$^{2}$ \\
    \textbf{Elias Stengel-Eskin}$^{3}$ \;
    \textbf{Hyunji Lee}$^{1}$ \;
    \textbf{Mohit Bansal}$^{1}$ \\
     \\
    University of North Carolina at Chapel Hill$^{1}$,
    Capital One$^{2}$,
    University of Texas at Austin$^{3}$\\
}

\begin{document}
\maketitle

\begin{abstract}
Various test-time interventions for Computer Use Agents (CUAs), including critic models, have been developed to improve performance
through pre-execution action evaluation in complex Graphical User Interface (GUI) environments.
However, existing critics suffer from two key limitations: they (1) focus primarily on \textit{short-sighted decision loops} (e.g., forgetting earlier actions) and (2) \textit{lack the visual grounding} needed to detect flawed actions (e.g., clicking wrong UI elements).
To address these, we introduce \ours, a \textbf{Hi}story-aware \textbf{Vis}ually \textbf{G}rounded test-time framework, built around a multimodal critic trained on real GUI trajectories to abstract past interactions into a compact record and to evaluate actions with visual grounding.
At test time, \ours integrates the critic into the policy decision loop to provide \textit{macro-action history}, which summarizes the policy's completed achievements, and \textit{visually grounded critique}, which verifies raw execution coordinates against the current screenshot to intercept errors before execution. 
Across web, mobile, and desktop benchmarks, \ours consistently outperforms existing scalar and verbal critics, improving average success rates over the strongest baseline by 5.8\% for Qwen3-VL-32B and 9.0\% for Gemini-3-Flash, and demonstrates strong cross-platform generalization.
Ablations show that macro-action history mitigates short-sighted planning and visually grounded critique reduces execution errors, with both components being critical for test-time scaling in long-horizon GUI tasks.\footnote{Code available at \href{https://github.com/G-JWLee/HiViG}{https://github.com/G-JWLee/HiViG}}
\end{abstract}

\begin{figure*}[t]
    \centering
    \includegraphics[width=\linewidth]{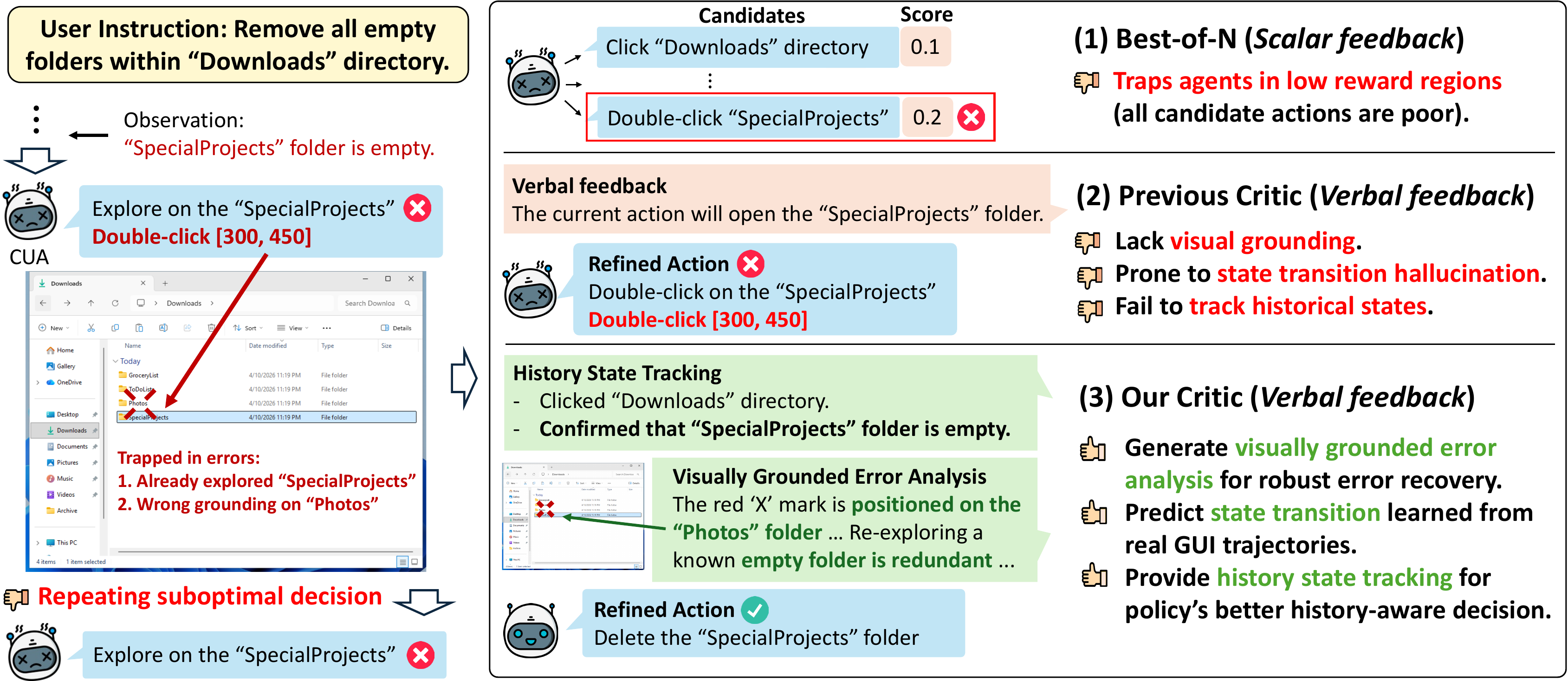}
    \par
    \vspace{-0.05in}
    \caption{
    Comparison of test-time interventions for Computer Use Agents (CUAs). \textbf{Left}: Lacking historical awareness and proactive error-recovery, standard policies easily become trapped in short-sighted decision loops. \textbf{Right}: Existing approaches are limited. Scalar feedback (top) traps policies in low-reward trajectory regions when all candidate actions are suboptimal. Previous critics (middle) rely heavily on textual intent, missing spatial and reasoning errors, and fail to provide historical awareness. In contrast, our critic (bottom) verifies raw execution coordinates, predicts immediate visual state outcomes grounded in its learned state transition knowledge, and provides visually grounded error analysis that intercepts errors before execution. Furthermore, it equips agents with history state tracking, condensing past interactions to guide them toward the final task objective. 
    }
    \vspace{-0.05in}
    \label{fig:teaser}
\end{figure*}

\section{Introduction}
Computer Use Agents (CUAs) are widely used to automate long-horizon tasks in Graphical User Interface (GUI) environments~\citep {Deng2023mind2web, He2024webvoyager, Koh2024visualwebarena}.
The underlying policy models inevitably make mistakes while operating directly on complex visual interfaces~\citep{Hong2024cogagent, Qin2025uitars, Awadallah2025fara7b}, such as selecting incorrect UI elements.
Since many GUI actions are irreversible~\citep{Wu2025guireflection}, relying on post-execution correction is often unsafe and impractical, motivating the use of test-time intervention to prevent errors before execution.
To provide such intervention, existing work employs reward models~\citep{Chae2025webshepherd, Xiong2025guipra, Zhang2026webarbiter} that offer \textit{scalar feedback} to score candidate actions.
However, in GUI environments with continuous parameter spaces (e.g., pixel coordinates), scoring alone is uninformative when all candidates are poor, trapping the policy in failure modes with no path to improvement~\citep{Luo2025languagemodelsverbalfeedback, Ning2026deaction}. 
In contrast, \textit{natural language (verbal) feedback} can explain why an action fails, enabling the policy to recover and progress toward task completion.
Thus, critic models that provide this kind of verbal feedback have emerged as a promising alternative to scalar reward models for test-time intervention.

Despite this conceptual shift, effective test-time critique, which serves as a pre-execution action evaluation to intercept a policy's proposed action before the action alters the environment, requires two underexplored capabilities by existing approaches. 
First, a reliable critic must maintain \textbf{visual grounding}.
Existing methods tend to rely on the policy's verbalized action intent (e.g., `double-click on ``SpecialProjects''' shown in~\Cref{fig:teaser} \highlight{(Right)}). 
As a result, they may erroneously approve a logically sound intent that is visually misaligned, such as an intent targeting incorrect coordinates or hallucinating state-transitions~\citep{Chae2025wma, Zheng2026code2world}. This risks allowing spatial or reasoning errors to bypass the pre-execution action evaluation.
To help a policy maintain task progress, a critic needs to keep track of \textbf{historical state} (i.e., what has been accomplished and failed).

\begin{figure*}[t]
    \centering
    \includegraphics[width=\linewidth]{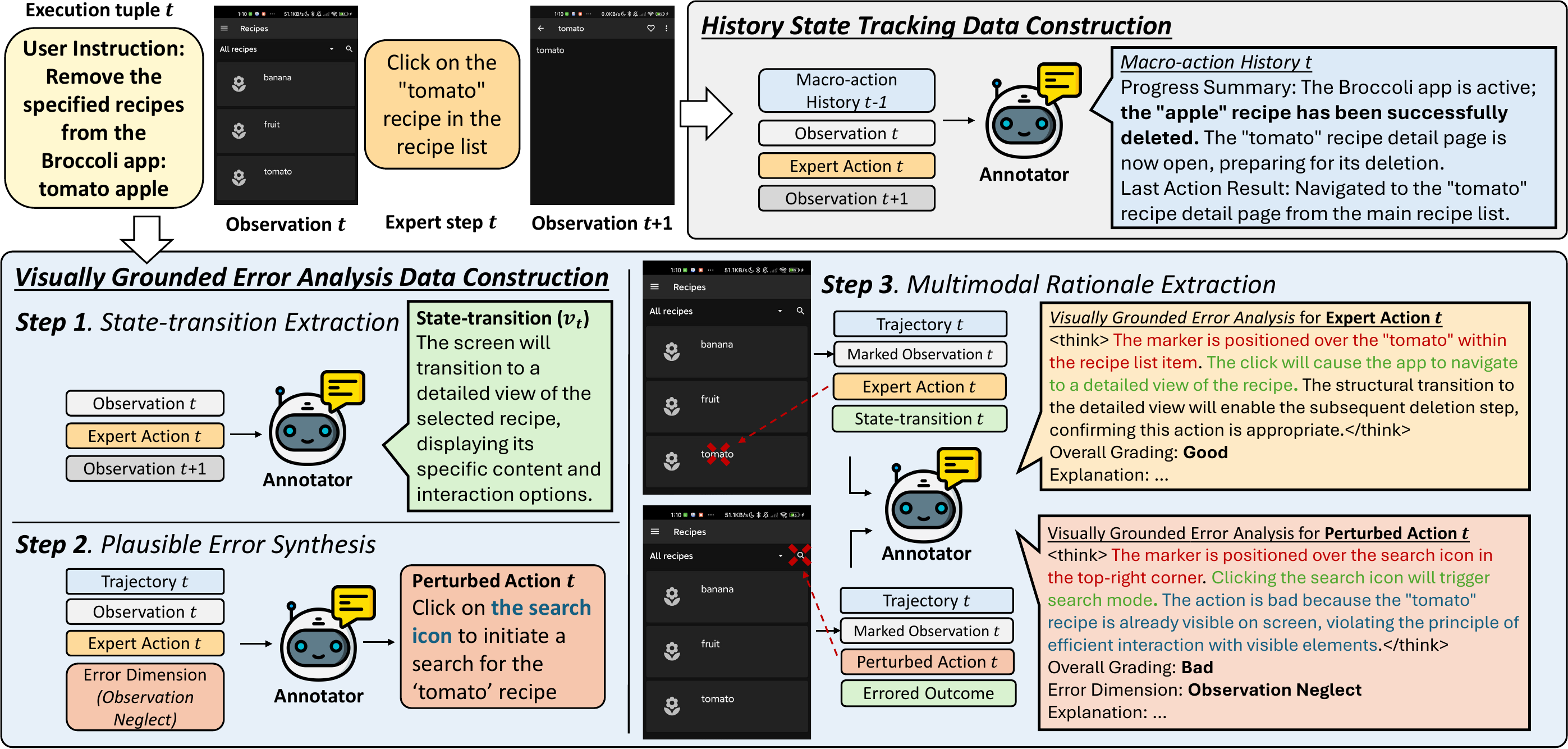}
    \par
    \vspace{-0.05in}
    \caption{Data construction for \ours. From the execution tuple (top left), we construct two distinct SFT datasets. \textbf{Top}: For history state tracking, the annotator iteratively translates visual state changes into a compact macro-action history to track long-term goal progress. \textbf{Bottom}: For visually grounded error analysis, we extract ground-truth state-transitions (Step 1) and synthesize plausible errors (Step 2). In Step 3, the annotator generates a multi-stage rationale that leverages a visual marker (red 'X') for visual grounding and the extracted state-transitions. (Note: User instructions are omitted here for simplicity.) 
    }
    \vspace{-0.05in}
    \label{fig:data_construction}
\end{figure*}

Our work aims to fill these gaps by introducing \textbf{Hi}story-aware \textbf{Vis}ually \textbf{G}rounded (\ours) test-time intervention framework~\Cref{fig:teaser} \highlight{(Right-bottom)}. 
\ours resolves prior shortcomings by equipping CUAs with two core capabilities.
First, \ours equips the policies with the ability to \textbf{track historical state}, enabling better history-aware planning over long horizons. To this end, \ours provides a macro-action history that summarizes past interactions to date.
By recursively compressing past interactions into multi-step achieved goals (e.g., ``Successfully opened the `Downloads' directory and confirmed the `SpecialProjects' folder is empty''), this helps the policy to track global task completion and avoid redundant decisions.
Second, to address the lack of visual grounding, \ours performs \textbf{visually grounded error analysis}.
Rather than overly relying on policy's textual intents, \ours verifies raw execution coordinates against actual visual states.
If a proposed action is flawed, our framework identifies the error dimension (e.g., visual hallucination, termination misjudgment) to provide the policy with corrective guidance before execution.
Inside our framework, we propose \ours-critic, a multimodal model trained to serve as a method for test-time intervention with these dual critique generation capabilities.
To this end, we construct a training corpus derived from open-sourced, multi-domain GUI trajectories~\citep{Liu2026scalecua}.
First, to teach history state tracking, we train the critic to update a macro-action history by integrating the last visual change with the past macro-action history to track long-term goal achievement (\Cref{fig:data_construction} \highlight{(Top-right)}). 
Second, to teach visually grounded error analysis, we train the critic to evaluate successful and flawed actions through a reasoning process: it verifies execution coordinates against the current screenshot for visual grounding, predicts the visual state-transition to assess the action's causal effect, and evaluates the action's relevance to the task (\Cref{fig:data_construction} \highlight{(Bottom-right)}).

\begin{figure*}[t]
    \centering
    \includegraphics[width=\linewidth]{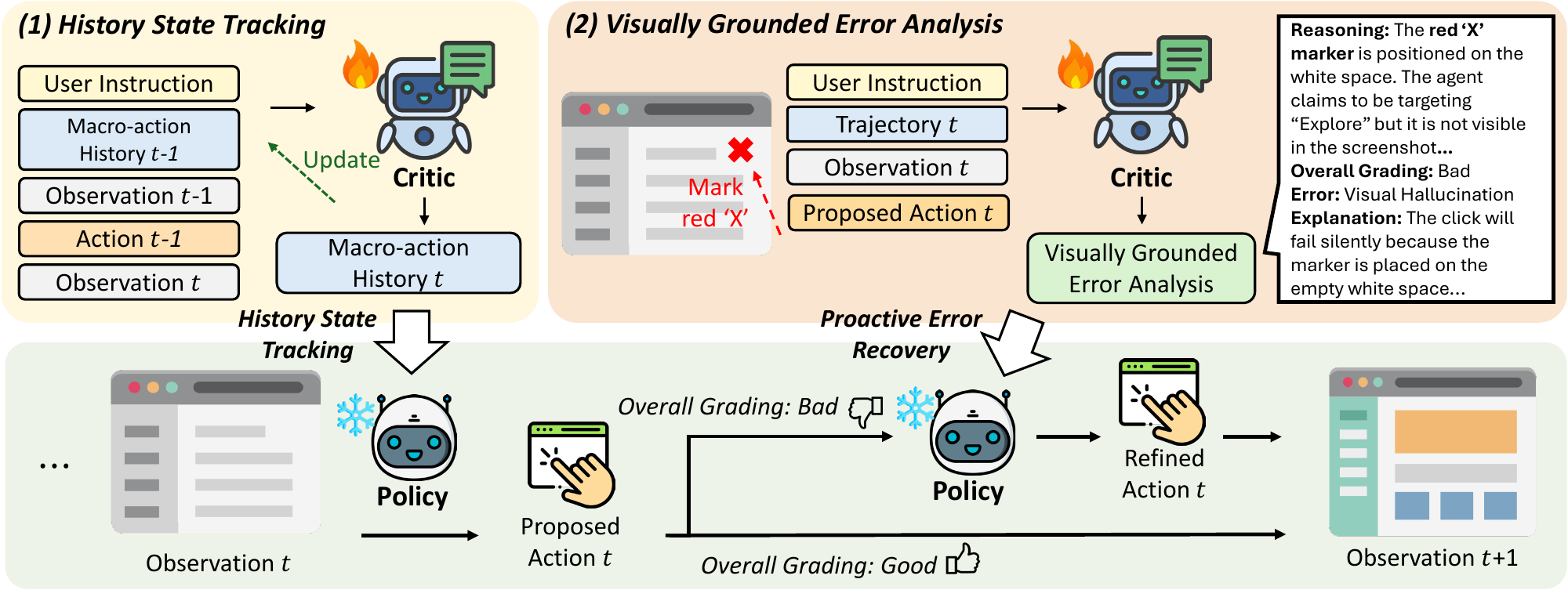}
    \par
    \vspace{-0.05in}
    \caption{Overview of \ours test-time intervention framework. \textbf{Top-left:} By analyzing visual changes between consecutive states, our critic updates the prior macro-action history, compressing micro-steps into macro-achievements. \textbf{Top-right:} our critic verifies raw pixel coordinates against the current visual state and predicts the impact of the proposed action, thereby catching spatial or reasoning errors before execution, outputting specific error dimensions to drive proactive error recovery.
    \textbf{Bottom:} At test time, the policy uses the macro-action history for better decision-making, and uses the visually grounded error analysis to refine the flawed action before execution. 
    }
    \vspace{-0.05in}
    \label{fig:overview}
\end{figure*}

To validate the efficacy of \ours, we conduct extensive evaluations across three diverse, long-horizon GUI benchmarks: WebArenaLitev2~\citep[web;][]{Liu2026scalecua}, AndroidLab~\citep[mobile;][]{Xu2025androidlab}, and WindowsAgentArena~\citep[desktop;][]{Bonatti2025windowsagentarena}. 
We evaluate our test-time intervention on two CUA policies: Qwen3-VL-32B-Thinking~\citep{qwen3-vl} and Gemini-3-Flash~\citep{gemini3}, representing open- and closed-weight models capable of navigating complex visual interfaces. 
Our empirical results~\Cref{tab:unified_performance} show that existing scalar rewards and visually ungrounded critics frequently degrade the performance of already highly-capable policies.
In contrast, \ours improves Gemini-3-Flash's overall success rate by 15.0\% (30.5\% to 45.5\%) on WebArenaLitev2 benchmark.
\ours also generalizes across different GUI environments, outperforming the strongest baseline critics by 9.0\% and 5.8\% when guiding Gemini-3-Flash and Qwen3-VL-32B-Thinking, respectively.
Overall, \ours is an test-time intervention framework that effectively guides strong policies to better complete long-horizon GUI tasks with history state tracking and visually grounded error analysis.

\section{\underline{Hi}story-aware \underline{Vis}ually \underline{G}rounded Test-time Intervention (\ours)}
\label{sec:method}
We first introduce the preliminaries of the CUA paradigm in \Cref{sec:sub:computer use agent preliminary}.
We then describe the data construction of two supervised fine-tuning (SFT) datasets jointly used to train the \ours-critic in \Cref{sec:sub:data construction} for \textit{history state tracking} and \textit{visually grounded error analysis}.
Finally, in \Cref{sec:sub:test-time guidance}, we present the overall \ours framework~(\Cref{fig:overview}) and describe how the trained critic is deployed at test time to guide policies.

\subsection{Preliminary: Computer Use Agent (CUA)}\label{sec:sub:computer use agent preliminary}
To automate digital tasks, we adopt the standard CUA paradigm, where a Multimodal Large Language Model (MLLM) operates as a policy that generates actions to interact with a Graphical User Interface (GUI) environment to fulfill a user instruction $I$~\citep{Wang2025opencua,Xu2026guiowl15}.
As illustrated in~\Cref{fig:overview} \highlight{(Bottom)}, at each timestep $t$, the environment provides a raw visual observation $o_t$ (i.e., a GUI screenshot).
The policy $\pi$ then generates a subsequent action $a_t$ conditioned on the instruction, the current observation $o_t$, and the execution history $H_{t-1}$, i.e., $a_{t}\sim\pi(\cdot|I,\;H_{t-1},\;o_t)$.
The trajectory $H_{t-1} = (a_{1},\; a_{2},\; \ldots,\; o_{t-W+1},\; a_{t-W+1},\; \ldots,\; o_{t-1},\; a_{t-1})$ is a structured history that, to manage MLLM context limitations in long-horizon GUI tasks, retains the complete sequence of past actions together with only the most recent $W$ visual observations~\citep{qwen3-vl, Wang2025opencua, Xu2026guiowl15}. 
The generated action $a_t$ typically consists of verbalized intent (e.g., `Click the browser's Back button in the top-left corner') and a pixel-based mouse or keyboard event (e.g., {\tt{\{``action'': ``left\_click'', ``coordinate'': [10, 20]\}}}).
The environment executes $a_t$ and transitions to the next visual state $o_{t+1}$, repeating until the policy generates a termination action or exhausts its computational budget.

\subsection{Critic Data Construction}\label{sec:sub:data construction}
We construct two distinct datasets to train \ours-critic for two capabilities: \textbf{(1) history state tracking} and \textbf{(2) visually grounded error analysis}, as shown in \Cref{fig:data_construction}. 
Specifically, we leverage an off-the-shelf MLLM as an annotator to automatically derive data from open-sourced, multi-domain GUI trajectories from ScaleCUA~\citep{Liu2026scalecua} training corpus.
By conditioning the annotator on the verified action execution outcomes present in the source corpus, we ensure the resulting data is grounded in environmental reality rather than the annotator's parametric reasoning.

\myparagraph{History State Tracking.}
To train our critic to track and supply the macro-action history, we construct a training dataset using an iterative compression mechanism (\Cref{fig:data_construction} \highlight{(Top-right)}).
First, given a successful trajectory, i.e., $(o_1,\;a_1,\;\ldots,\;o_t,\;a_t)$ from the source corpus, we segment the sequence into overlapping execution tuples consisting of the user instruction, the previous visual observation, the executed action, and the resulting visual observation: $\tau_{t}=(I,\;o_{t},\;a_{t},\;o_{t+1})$, as shown in~\Cref{fig:data_construction} \highlight{(Top-left)}.
At timestep $t$, the MLLM annotator takes the previous macro-action history $H_{t-1}$ and the tuple $\tau_t$ as inputs.
The annotator translates the visual difference between the two consecutive observations into text to identify valid UI updates or silent failures. 
Merging this analysis with the previous macro-action history $H_{t-1}$, it recursively compresses past atomic steps into an updated macro-action history $H_{t}$ containing multi-step achieved goals, which our critic leverages to guide the policy to have better history state understanding during its next-step planning. 

\myparagraph{Visually Grounded Error Analysis.}
To enable our critic to penalize errors from policy actions, we construct a dataset that captures diverse failure modes using visually grounded rationales. 
The dataset construction process consists of \textit{three} steps: (i) state-transition extraction, (ii) plausible error synthesis, and (iii) multimodal rationale extraction.

\noindent\textbf{{Step 1. State-transition Extraction.}}
To ground error analysis on real GUI state changes, we represent the immediate visual impact of each successful action in natural language~\citep{Chae2025wma,Mei2025rwom}.
Given the execution tuple $\tau_{t}$, we prompt the MLLM annotator to generate a verbalized state-transition $v_t$ that describes the observed causal effect of the action.
Because the annotator conditions on the actual resulting screenshot $o_{t+1}$, $v_t$ reflects a visual change rather than inferred intent (\Cref{fig:data_construction} \highlight{(Bottom-left)}).
These verbalized transitions provide a compact and interpretable representation of short-term dynamics, acting as the basis for rationale generation in \textbf{Step 3}.

\noindent\textbf{{Step 2. Plausible Error Synthesis.}} 
To synthesize a comprehensive set of common GUI failure modes reported in recent CUA literature~\citep{Li2025screenspotpro, Wang2025opencua, Liu2026scalecua, Jin2026halluclear}, we prompt the annotator to systematically perturb the \textit{expert} action $a_{t}$ into a \textit{perturbed} action $\hat{a}_{t}$ (\Cref{fig:data_construction} \highlight{(Bottom-left)}).
This generation is guided by a taxonomy of 12 diverse error dimensions observed in CUA literature (the complete taxonomy, the distribution of each error type, and representative examples are in~\Cref{appendix:Error Dimensions}), including \textit{Grounding Errors} (e.g., intent is correct but clicking adjacent non-interactive space), \textit{Procedural Prerequisite Neglect} (e.g., typing without focusing the search field), and \textit{Visual Hallucinations} (e.g., targeting a non-existent UI element).
Conditioned on the user instruction $I$, past interaction trajectory $h_t=(a_{1},\;a_{2},\;\ldots,\;a_{t-1})$, current observation $o_t$, and expert action $a_{t}$, the annotator selects the most appropriate error from the taxonomy to ensure these synthesized errors are plausible. 
To prevent hallucinations and guarantee the quality of these synthesized actions, we employ an automated double-blind filtering pipeline using a multimodal judge (Qwen3-VL-32B). A human validation study confirms our filtering protocol highly correlates with human consensus (76.6\% agreement); we detail this study in Appendix~\ref{appendix:human_validation}.

\noindent\textbf{{Step 3. Multimodal Rationale Extraction.}}
Using the expert action $a_{t}$ and perturbed action $\hat{a}_{t}$, we prompt the annotator to generate fine-grained, step-by-step rationales, as shown in~\Cref{fig:data_construction} \highlight{(Bottom-right)}.
To enforce spatial awareness lacking in prior critics, we apply two strategies 
during data construction.
First, to break the model's over-reliance on text, we mask the policy's verbal intent in $30\%$ of samples.
Second, we render a visual marker~\citep{Yang2023som} on the screenshot $o_t$ at the exact proposed coordinates and instruct the annotator to reason based on the marker, extracting the supervisory signal needed to teach the target critic to perform action verification using this marker position at test time. 
Conditioned on the provided action causal effect, which is either the ground-truth state-transition $v_{t}$ for $a_{t}$ or the pre-defined failure outcome for $\hat{a}_{t}$ (e.g., clicking inactive space yields no UI update in Ground Error), and this marked screenshot, the annotator structures a rigorous reasoning process. 
Specifically, it first visually verifies the action by identifying the UI element directly beneath the marker, preventing it from overly relying on the textual intent.
Next, incorporating this visual verification and provided action causal effect, the annotator evaluates the action's alignment with the user instruction $I$, and outputs verbal feedback that can include a corresponding error dimension and a corrective explanation to guide a policy.
This pipeline yields diverse SFT samples of visually grounded rationales (\Cref{fig:data_construction} \highlight{(Right-bottom)}), providing the optimal data to teach our critic precise action verification.

\subsection{Test-time CUA Guidance}\label{sec:sub:test-time guidance}
During deployment, \ours framework operates alongside the underlying CUA policy and \ours-critic provides continual history tracking and error recovery.
As illustrated in~\Cref{fig:overview}, our framework intervenes in two distinct phases at each timestep $t$.
First, before the policy plans its next move, the critic updates the macro-action history.
By processing the previous execution tuple $\tau_{t-1}=(I,\;o_{t-1},\;a_{t-1},\;o_t)$ and prior history $H_{t-1}$, it produces an updated history $H_t$.
By articulating visual changes and mapping them to global task progress, macro-action history equips the policy with the history state tracking required to propose an informed initial action $a_t$.
Once proposed, the critic verifies this action.
To ground the critic's analysis in visual observation, we render the visual marker at the action's proposed coordinates on the screenshot.
Given the user instruction $I$, the execution trajectory $h_t$, and this marked observation, the critic evaluates the proposed action using its learned reasoning process: visual verification, state-transition prediction, and instruction alignment.
If graded ``Good'', the action is directly executed, advancing the GUI to the next observation $o_{t+1}$.
If the action is rated ``Bad'', the critic classifies the failure into one of the predefined errors and generates a verbal explanation.
This pre-execution evaluation serves as an constraint, forcing the policy to refine its action, and enables continuous progress toward the goal, preventing the compounding of silent failures.

\section{Experiments}
\label{experiments}

\subsection{Setup}
\myparagraph{Computer Use Agents.}
To execute complex GUI tasks and evaluate test-time intervention frameworks, we use two distinct Multimodal Large Language Models (MLLMs) as Computer Use Agent (CUA) policies: Qwen3-VL-32B-Thinking~\citep{qwen3-vl}, an open-source model; and Gemini-3-Flash~\citep{gemini3}, a closed-source frontier model.
We employ a generalized pixel-based action space adapted from \citet{qwen3-vl}, where policies interact with environments by generating structured JSON tool calls encompassing standard operations (e.g., click, type, scroll, swipe). 
The complete specifications for the desktop and mobile action spaces can be found in~\Cref{appendix:action_space}.

\begin{table*}[t]
    \small
    \centering
        \renewcommand{\arraystretch}{1.05}
        \setlength{\tabcolsep}{12pt}
        \begin{tabular}{l c c c c c c c c}
             \toprule
             {\multirow{2}{*}{\textbf{Method}}} & \multicolumn{4}{c}{\textit{Qwen3-VL-32B-Thinking}} & \multicolumn{4}{c}{\textit{Gemini-3-Flash}}\\
             \cmidrule(l{0.5em}r{0.5em}){2-5} \cmidrule(l{0.5em}r{0.5em}){6-9}
              & {\textbf{WALv2}} & {\textbf{ALab}} & {\textbf{WAA}} & {\textbf{Avg.}} & {\textbf{WALv2}} & {\textbf{ALab}} & {\textbf{WAA}} & {\textbf{Avg.}} \\
             \midrule
             Base agent &
             {\small 13.0} & {\small 44.2} & {\small \underline{35.7}} & {\small 31.0} & {\small \underline{30.5}} & {\small \underline{58.0}} & {\small 35.8} & {\small \underline{41.4}} \\
             \cmidrule{1-9}
             OpenCUA &
             {\small 14.9} & {\small 47.1} & {\small 35.2} & {\small 32.4} & {\small 29.9} & {\small 57.2} & {\small 35.8} & {\small 41.0} \\
             SE-WSM &
             {\small 11.7} & {\small 45.7} & {\small 31.6} &  {\small 29.7} & {\small \underline{30.5}} & {\small 57.2} & {\small 35.1} & {\small 40.9} \\
              \cmidrule{1-9}
             CGI (32B) &
             {\small \underline{16.9}} & {\small 46.9} & {\small 33.7} & {\small \underline{32.5}} & {\small 29.9} & {\small 55.8} & {\small \underline{37.9}} & {\small 41.2} \\
             CGI (8B) &
             {\small 12.3} & {\small 44.2} & {\small 31.0} & {\small 29.2} & {\small 26.6} & {\small 55.1} & {\small \underline{37.9}} & {\small 39.9} \\
             GUI-Critic-R1 &
             {\small 13.6} & {\small \underline{49.3}} & {\small 34.4} & {\small 32.4} & {\small 22.1} & {\small 53.6} & {\small 32.5} & {\small 36.1} \\
             \cmidrule{1-9}
             \cellcolor{gg}\ours (Ours) &
             \cellcolor{gg}{\small \textbf{25.3}} & \cellcolor{gg}{\small \textbf{51.5}} & \cellcolor{gg}{\small \textbf{38.0}} & \cellcolor{gg}{\small \textbf{38.3}} & \cellcolor{gg}{\small \textbf{45.5}} & \cellcolor{gg}{\small \textbf{61.6}} & \cellcolor{gg}{\small \textbf{44.2}} & \cellcolor{gg}{\small \textbf{50.4}} \\
             \bottomrule
        \end{tabular}
    \vspace{-0.05in}
    \caption{Comparison of test-time interventions on Qwen3-VL-32B and Gemini-3-Flash across GUI benchmarks in diverse environments. The best and the second best overall success rate results are in \textbf{bold} and \underline{underline}, respectively. \ours delivers gains in all benchmarks. WALv2: WebArenaLitev2 for web, ALab: AndroidLab for mobile, WAA: WindowsAgentArena for desktop, and Avg.: average overall success rate. Detailed performances for each benchmark are in~\Cref{tab:webarenalitev2}, \Cref{tab:androidlab}, and~\Cref{tab:windowsagentarena}.
    }
    \label{tab:unified_performance}
    \vspace{-0.05in}
\end{table*}

\myparagraph{Dataset.}
To train our critic, we construct SFT dataset using the ScaleCUA~\citep{Liu2026scalecua} training corpus which contains GUI trajectories spanning web, mobile, and desktop environments.
Using Qwen3-VL-32B-Thinking as the MLLM annotator, we generate 52k mixed samples, including 20k history state tracking and 32k visually grounded error analysis.

For evaluation, we conduct experiments across three representative computing ecosystems: web, mobile, and desktop.
For the web environment, we use WebArenaLitev2~\citep{Liu2026scalecua, Zhou2024webarena}, comprising vision-centric tasks performed in 5 diverse websites with a 15-step maximum horizon.
To assess mobile GUI control, we employ AndroidLab~\citep{Xu2025androidlab}, evaluating agents across 9 mobile applications with task-specific step limits.
Finally, for the desktop environment, we use WindowsAgentArena~\citep{Bonatti2025windowsagentarena}, comprising Windows 11 OS tasks grouped into 6 categories~\citep{Rivard2025neuralos} bounded by a 30-step horizon.
Across all benchmarks, performance is measured by execution-based task success rates. Further details are in~\Cref{appendix:experiment setup}.

\myparagraph{Baselines.}
We compare our approach against three distinct categories of test-time intervention: 
\textbf{(1) Base agent:} the unguided CUA, serving as a reference that executes the policy without any test-time intervention.
\textbf{(2) Scalar feedback (PRMs):} To compare against standard Process Reward Models (PRMs) operating in a Best-of-N search, we adapt established step-evaluation templates~\citep{Lin2025cuarewardbench} from OpenCUA~\citep{Wang2025opencua} and SE-WSM~\citep{Sun2025sewsm}. We simplify these into standard PRM templates using the same backbone model (Qwen3-VL-8B-Thinking) as our critic, and set $N=2$ to match the computational cost of verbal frameworks, which allow a maximum of two policy calls per step.
\textbf{(3) Verbal feedback:} To benchmark against existing critics, we evaluate two base models (Qwen3-VL-8B-Thinking and Qwen3-VL-32B-Thinking) as a critic prompted with the zero-shot multidimensional critique prompt from CGI~\citep{Yang2025cgi}. 
We also compare against GUI-Critic-R1~\citep{Wanyan2025guicriticr1}, a specialized critic model trained for mobile GUI environments.  
Further details regarding prompt configurations and baseline implementations are provided in~\Cref{appendix:experiment setup}.

\myparagraph{Implementation Details.}
We initialize our critic with Qwen3-VL-8B-Thinking~\citep{qwen3-vl} base model, which is trained for one epoch on the mixed SFT dataset. Further details regarding training hyperparameters, computing resources, and inference configurations are provided in~\Cref{appendix:experiment setup}.

\subsection{Results and Discussion\label{sec:experiments:exp_results}}

\myparagraph{\ours outperforms existing test-time interventions across diverse GUI platforms and policies.} 
To evaluate the efficacy and cross-platform generalizability of \ours, we benchmark test-time interventions across three distinct GUI environments: web (WebArenaLitev2), mobile (AndroidLab), and desktop (WindowsAgentArena).
As shown in~\Cref{tab:unified_performance}, \ours consistently improves both the open-weight Qwen3-VL-32B-Thinking and the frontier closed-source Gemini-3-Flash policies, achieving average absolute gains in overall success rate of 7.3\% and 9.0\% across the three domains, respectively.
These improvements are observed in all three environments, demonstrating strong cross-platform generalization. This strong generalization stems from \ours relying solely on raw GUI screenshots and pixel-level visual understanding, without requiring platform-specific features such as DOM or accessibility tree.

\ours consistently achieves performance gains over the Qwen3-VL-32B-Thinking as the base policy, outperforming strong baselines.
Specifically, our framework surpasses the base agent by 12.3\%, 7.3\%, and 2.3\% in absolute overall success rate on WebArenaLitev2, AndroidLab, and WindowsAgentArena, respectively, and surpasses the best baseline on each benchmark by 5.8\% average overall success rate. 
We additionally evaluate \ours using the highly optimized, frontier-class Gemini-3-Flash as the policy backbone, to demonstrate the versatility of our test-time intervention approach.
Although Gemini-3-Flash is already strong, \ours improves its absolute success rate by 15.0\%, 3.6\%, and 8.4\% on WebArenaLitev2, AndroidLab, and WindowsAgentArena, respectively.
These results suggest that limited history state tracking and ineffective error recovery remain key challenges for CUAs, even when powered by frontier models.

We observe that \ours is particularly beneficial for challenging tasks where existing test-time interventions often fail. 
Without an abstractive macro-action history, policies frequently repeat ineffective commands, as they lack a compressed record of prior attempts that would allow them to avoid short-sighted decision loops.
Without visually grounded critiques to verify raw execution coordinates, critics often fail to detect misaligned actions caused by complex UI elements.
In contrast, \ours mitigates these failures by maintaining a macro-action history and providing visually grounded critiques, allowing policies to make decisions anchored in visual content and historical progress.
As a result, it improves performance on tasks that otherwise remain very difficult, increasing Qwen3-VL-32B-Thinking's success rate on the WebArenaLitev2 Map category from 3.9\% to 23.1\% (\Cref{tab:webarenalitev2}), and Gemini-3-Flash's success rate on the WindowsAgentArena Office category from 4.7\% to 23.3\% (\Cref{tab:windowsagentarena}).

\myparagraph{Lack of visual grounding and history conditioning hurts other critics.}
In~\Cref{tab:unified_performance}, compared to our approach, baseline interventions often degrade frontier model performance, yielding at best a 2.1\% gain on WindowsAgentArena and at worst an 8.4\% drop on WebArenaLitev2 for Gemini-3-Flash.
Notably, scalar feedback baselines (OpenCUA and SE-WSM) yield no performance improvement over the base agent.
This shows a fundamental limit of scalar rewards: when all candidate actions lead to silent failures, scalar scores offer no guidance to improvement.
Furthermore, while verbal feedback approaches (CGI, GUI-Critic-R1) offer more expressive signals, their improvements are highly inconsistent across domains.
For instance, GUI-Critic-R1 achieves a 5.1\% absolute overall success rate gain on AndroidLab, but only a marginal 0.6\% on WebArenaLitev2, and degrades performance by 1.3\% on WindowsAgentArena.
This instability stems from two fundamental limitations: 
First, existing critics lack the visual grounding (see a qualitative example in~\Cref{fig:visual_grounding}) needed to generalize across the diverse, dense layouts of web and desktop interfaces.
Second, they fail to track historical state, limiting the policy on tasks requiring long execution trajectories (examples in~\Cref{fig:teaser} and~\Cref{fig:history_state_tracking}). 
Further analysis on the lack of visual grounding and history state tracking of baseline critics are in~\Cref{appendix:sub:Lack of Visual Grounding in Previous Critics} and ~\Cref{appendix:sub:Lack of History State Tracking in Previous Critics}.

\begin{table}[t]
    \centering
    \begin{minipage}{0.48\textwidth}
    \centering
    \small
        \renewcommand{\arraystretch}{1.3}
        \setlength{\tabcolsep}{5.0pt}
        \begin{tabular}{c c c c}
             \toprule
             {\textbf{VisAnalysis}} & {{\textbf{HisTrack}}} & \textit{Qwen3-VL} & \textit{G-3-Flash}\\
             \midrule
             \xmark & \xmark & {\small 13.5} & {\small 30.0} \\
             \cmark & \xmark & {\small 21.4} & {\small 35.1} \\
             \xmark & \cmark & {\small 23.4} & {\small 42.9} \\
             \cmark & \cmark & {\small \textbf{25.3}} & {\small \textbf{45.5}} \\
             \bottomrule
        \end{tabular}
    \vspace{-0.05in}
    \caption{\textbf{Impact of verbal feedback components.} Experiments with components of \ours on WebArenaLitev2 evaluated with Qwen3-VL-32B and Gemini-3-Flash as CUA. VisAnalysis: Visually-grounded error analysis, HisAnalysis: History state tracking. Combining two components leads to better performance. 
    }
    \label{tab:component_ablation}
    \end{minipage}
    \vspace{-0.05in}
\end{table}

\subsection{Analysis and Ablations\label{sec:experiments:ablations}}

\begin{table*}[t]
    \centering
    \begin{minipage}{\textwidth}
    \centering
    \small
        \renewcommand{\arraystretch}{1.3}
        \setlength{\tabcolsep}{8.pt}
        \begin{tabular}{l c c c c}
             \toprule
             \textbf{Method} & \textbf{Policy Inference} & \textbf{Critic Inference} & \textbf{Latency (s)} & \textbf{WALv2 Avg. (\%)} \\
             \midrule
             Base agent & 1 & 0 & \textbf{8.2} & 13.0 \\
             OpenCUA & 2 & 2 & 31.8 & 14.9 \\
             SE-WSM & 2 & 2 & 41.7 & 11.7 \\
             CGI (32B) & 1-2 & 1 & 37.9 & 16.9 \\
             CGI (8B) & 1-2 & 1 & 21.5 & 12.3 \\
             GUI-Critic-R1 & 1-2 & 1 & 14.8 & 13.6 \\
             \midrule
             VisAnalysis (Ours) & 1-2 & 1 & 15.2 & 21.4 \\
             HisAnalysis (Ours) & 1 & 1 & 11.5 & 23.4 \\
             \ours (Ours) & 1-2 & 2 & 17.4 & \textbf{25.3} \\
             \bottomrule
        \end{tabular}
    \vspace{-0.05in}
    \caption{\textbf{Performance-latency trade-off.} Average step latency (wall-clock time) and the average success rate on WebArenaLitev2. Policy Inference and Critic Inference indicate the number of policy and critic model inferences per step, respectively, at test-time. Latency is measured across 20 tasks using 4 H100 GPUs. \ours demonstrates superior performance while maintaining a highly competitive runtime. VisAnalysis: Visually-grounded error analysis, HisAnalysis: History state tracking. Combining two components leads to better performance. 
    }
    \label{tab:latency_tradeoff}
    \end{minipage}
    \vspace{-0.15in}
\end{table*}
\begin{table}[t]
    \centering
    \begin{minipage}{0.48\textwidth}
    \centering
    \small
        \renewcommand{\arraystretch}{1.3}
        \setlength{\tabcolsep}{5.pt}
        \begin{tabular}{c c c c}
             \toprule
             {\textbf{Intent Masking}} & {{\textbf{Visual Marker}}} & WALv2 & ALab \\
             \midrule
             \cmark & \xmark & {\small 20.8} & {\small 46.4} \\
             \xmark & \cmark & {\small \textbf{25.3}} & {\small 47.1} \\
             \cmark & \cmark & {\small \textbf{25.3}} & {\small \textbf{51.5}} \\
             \bottomrule
        \end{tabular}
    \vspace{-0.05in}
    \caption{\textbf{Ablation of visual grounding strategies.} We evaluate the impact of masking the policy's verbal intent (30\% of SFT samples) and injecting a visual marker (i.e., 'X' marker). Combining both strategies forces the critic to break its text reliance and evaluate the raw spatial coordinates, yielding visually accurate verbal feedback. WALv2: WebArenaLitev2, ALab: AndroidLab.
    }
    \label{tab:visual_grounding_ablation}
    \end{minipage}
    \vspace{-0.05in}
\end{table}

\myparagraph{Effectiveness of Individual Components in \ours Framework.}
To understand the individual contributions of visually grounded error analysis and history state tracking in our framework, we ablate these components on WebArenaLitev2, as shown in~\Cref{tab:component_ablation}.
Deploying either component alone outperforms all baselines.
This is especially notable with the highly-capable Gemini-3-Flash policy.
While prior baselines degrade its performance on WebArenaLitev2 (\Cref{tab:unified_performance}), applying our critic solely with visually grounded error analysis increases its success rate from 30.5\% to 35.1\%.
Furthermore, deploying history state tracking alone achieves 23.4\% and 42.9\% overall success rates for Qwen3-VL-32B and Gemini-3-Flash, respectively.
By tracking the task progress, this component prevents the policy from falling into short-sighted decision loops and needlessly re-exploring known states.
This independent strength also shows the computational flexibility of our framework; while the full pipeline requires two critic inferences per step, a single inference of either capability is already sufficient to achieve better results. 
Combining these two within \ours yields the highest overall success rates (25.3\% for Qwen3-VL-32B-Thinking and 45.5\% for Gemini-3-Flash), proving a strong synergistic effect of the two components that are essential to \ours's superiority. More detailed analyses of these components are provided in~\Cref{appendix:sub:impact of mixed dataset} and~\Cref{appendix:sub:synergistic effects of two components}.

\myparagraph{Importance of Visual Grounding.}
Prior critics often struggle to enforce visual grounding effectively in their training mechanisms and erroneously approve spatially misaligned actions by over-relying on the policy's verbal intent.
To overcome this, we enforce visual grounding during Multimodal Rationale Extraction (\textbf{Step 3}) in~\Cref{sec:sub:data construction}: masking the verbal intent and injecting a visual marker on the screenshot.
\Cref{tab:visual_grounding_ablation} demonstrates the efficacy of combining both strategies.
Training without intent masking degrades AndroidLab performance from 51.5\% to 47.1\%.
This exposes a ``shortcut'' phenomenon; if verbal intent is consistently available during training, the model ignores the visual observation, missing critical spatial information.
When trained without the visual marker, performance drops significantly (e.g., from 25.3\% to 20.8\% on WebArenaLitev2). 
This indicates that MLLMs struggle to interpret raw numerical coordinates in complex visual interfaces, and the marker can act as a reliable and effective spatial anchor to help comprehend the visual semantics of the action.
Overall, having both strategies shows the best performance, where the intent masking breaks text-reliance to force analysis on the screenshot, while the visual marker provides spatial anchors needed to verify action execution.
Additionally, we provide the analysis of the importance of state-transition knowledge in~\Cref{appendix:sub:impact of state-transition}.

\myparagraph{Performance and Latency Trade-off.}
To better understand the computational efficiency of \ours test-time intervention, we evaluate explicit wall-clock time against baseline methods.
We measure the average step latency across 20 tasks in WebArenaLitev2 using 4 H100 GPUs, allocating 2 GPUs for the Qwen3-VL-32B policy and 2 for the critic models.
As detailed in~\Cref{tab:latency_tradeoff}, \ours allows for flexible trade-offs between accuracy and efficiency. 
Employing either the visually grounded error analysis or the history state tracking component alone requires only a single critic inference, achieving average latencies of 15.2s and 11.5s, respectively. 
This shows that \ours's individual components are as computationally efficient as GUI-Critic-R1 (14.8s) while significantly outperforming it (21.4\% and 23.4\% vs.\ 13.6\%). 
Furthermore, deploying the full \ours pipeline with two critic inferences requires 17.4s. 
This remains substantially faster than 32B-scale critics like CGI (37.9s) while delivering far superior performance (25.3\% vs.\ 16.9\%). 
Overall, \ours introduces a latency increase over the base agent (8.2s), but achieves the highest performance among all evaluated methods in~\Cref{tab:latency_tradeoff}.

\section{Related work}
\label{related_work}

\myparagraph{Reward Models for Scalar Feedback.}
To mitigate compounding errors in long-horizon GUI tasks, recent test-time interventions employ reward models to score candidate actions~\citep{Xu2025raggui, Cheng2025atlas, Mei2025rwom, Chen2025guishepherd}.
World-model-style reward models~\citep{Chae2025wma, Bai2025digiq} predict action outcomes and assign scalar scores to future states.
However, reliably generating precise future states in complex GUIs is challenging~\citep{Cheng2025atlas, Zheng2026code2world}.
Moreover, evaluating these imperfect simulations with outcome-based scalar rewards risks compounding errors and provides no path to improvement when all candidate actions are poor.
While some work augments reward models with external tutorials~\citep{Xu2025raggui, Mei2025rwom} to improve planning, this reliance can limit generalizability,
as such resources are rarely available for proprietary or diverse GUI environments.
In contrast, \ours utilizes a computationally efficient 8B model and generalizes across diverse visual interfaces by relying purely on screenshot-level visual understanding of the current state.

\myparagraph{Verbal Feedback Critics.}
In general, scalar feedback offers little guidance to the policy when all candidate actions are poor. 
To address this, verbal critics~\citep{Xiong2026phycritic} provide natural-language critiques~\citep{Luo2025languagemodelsverbalfeedback, Zhong2024languagefeedbackmodels} that help policies refine their trajectories~\citep{Wu2025guireflection, Tang2025refcritic, Yang2025cgi}.
This refinement process involves executing an action, observing the environment outcome, and backtracking to correct mistakes.
However, generating critiques after an action is executed is highly impractical in real-world GUIs, as many actions are irreversible (e.g., sending an email or deleting a file).
GUI-Critic-R1~\citep{Wanyan2025guicriticr1} addresses this by extending critique generation to the pre-execution setting, but lacks the visual grounding 
to verify spatial correctness and does not explicitly reason over historical interactions.
Similarly, OS-Oracle~\citep{Wu2025osoracle} proposes a framework 
for cross-platform GUI critics using taxonomy-guided error synthesis.
However, OS-Oracle relies primarily on prompting frontier models (e.g., GPT-4o) for rationale generation, lacking strict mechanisms to enforce pixel-level spatial verification or explicit history state tracking.
In contrast, \ours provides history-aware, visually grounded critiques by predicting the visual consequence of a proposed action from the current visual observation and grounding its assessment in past trajectories, enabling safer and more reliable test-time decision making. 

\section{Conclusion}
In this paper, we introduce \ours, a test-time intervention framework that equips CUAs with history state tracking and visually grounded error analysis.
\ours leverages a multimodal critic trained to summarize past interactions that enables policies to track long-term progress and a visually grounded critique that intercepts spatial errors before execution by verifying raw execution coordinates against the screenshot.
Evaluations show \ours generalizes across web, mobile, and desktop environments, consistently enhancing frontier models and achieving average overall success rates by up to 9.0\%.
Our ablations confirm a strong synergy between the two components, proving that history-aware and visually grounded test-time intervention can push the performance limits of frontier policies without modifying their underlying weights.

\section*{Limitations}
While our test-time intervention framework shows strong performance across diverse policies and GUI environments, our taxonomy remains bounded by the current landscape of digital interfaces.
Although our 12 predefined dimensions capture the most prevalent GUI failure modes, the continuous evolution of digital interfaces and policy capabilities will likely give rise to new classes of spatial and reasoning errors, necessitating an iterative expansion of the existing taxonomy. 

\section*{Acknowledgments}
This work was supported by NSF-AI Engage Institute DRL2112635, NSF-CAREER Award 1846185, ARO Award W911NF2110220, ONR Grant N00014-23-1-2356, Capital One Research Award, Apple PhD Fellowship, NDSEG PhD Fellowship. The views contained in this article are those of the authors and not of the funding agency.

\bibliography{custom}

\clearpage
\appendix
\crefalias{section}{appendix}

\section{Details of Experimental Setups\label{appendix:experiment setup}}
\myparagraph{Training Dataset.}
To construct the SFT datasets for our critic, we use the ScaleCUA~\citep{Liu2026scalecua} training corpus as source data that contains multi-domain GUI trajectories, including web, mobile, and desktop. 
We employ Qwen3-VL-32B-Thinking as the MLLM annotator to extract the multimodal rationale for error analysis and generate the macro-action history.
The resulting training data contains 52k mixed SFT samples: 20k samples for the history state tracking task, and 32k samples for the visually grounded error analysis task, containing 16k samples for expert actions and 16k samples for perturbed actions.

\myparagraph{Evaluation Benchmarks.}
We provide in-depth explanations of the cross-platform GUI navigation benchmarks used in our experiments, following the evaluation setups in~\citet{Liu2026scalecua}.
For the web environment, WebArenaLitev2~\citep{Liu2026scalecua} serves as a vision-centric adaptation of the WebArena~\citep{Zhou2024webarena} framework, consisting of 154 tasks that span diverse interactive websites, including online shopping, content management platforms, map services, GitLab, and Reddit. 
To assess mobile GUI control, AndroidLab~\citep{Xu2025androidlab} utilizes a pixel-grounded visual interaction mode to test true spatial understanding, evaluating agents across 138 complex tasks distributed over 9 native applications such as the calendar, clock, contacts, map, and settings. 
Finally, for the desktop environment, WindowsAgentArena~\citep{Bonatti2025windowsagentarena} tests agents within a fully functional Windows 11 virtual machine across 145 diverse tasks. 
We chose a 30-step execution horizon as an intermediate setting between the shorter 15-step horizons and the 50-step limits in the previous paper~\citep{Liu2026scalecua, Yang2026ossymphony}, which allows us to evaluate the policy's ability to solve long-horizon tasks without high costs.
To thoroughly evaluate performance across varying operational complexities, we follow the functional categorization in~\citet{Yang2026ossymphony}, grouping these tasks into six specific Windows OS application scenarios: Office, Web Browsing, Windows System, Code, Media, and Windows Utilities.
ScaleCUA and WebArenaLiteV2 are released under the Apache 2.0 License, and AndroidLab and WindowsAgentArena are released under the MIT License.

\myparagraph{Baselines.}
We provide a more detailed explanation of the baseline models and their specific prompt configurations used in our experiments.

\begin{itemize}[itemsep=2mm, parsep=1pt, leftmargin=*]
\item \textbf{OpenCUA}~\citep{Wang2025opencua} originally employs a strong proprietary model (e.g., Claude) as a step reflector in their chain-of-thought data annotation pipeline. It generates reflections based on previous step reasoning and current screenshots. To avoid coupling complexities and ensure a fair comparative evaluation, we simplify their reflector prompt into a standard Process Reward Model (PRM) template. Given the prompt, Qwen3-VL-8B-Thinking base model outputs a scalar quality score. 

\item \textbf{SE-WSM}~\citep{Sun2025sewsm} conducts a comprehensive, step-by-step analysis of input trajectories, providing multidimensional evaluations that cover trajectory correctness, the identification of redundant steps, the first error step, and correct action suggestions. We adapt SE-WSM's dense prompt template for our scalar feedback baseline to evaluate candidate actions in a Best-of-N search, also utilizing Qwen3-VL-8B-Thinking as the base evaluator.

\item \textbf{CGI}~\citep{Yang2025cgi} is a zero-shot critique generation framework. We implement this by prompting our base models (Qwen3-VL-8B-Thinking and Qwen3-VL-32B-Thinking) to systematically analyze the proposed action across three explicit dimensions: \textit{Contribution} (task progress), \textit{Feasibility} (validity within the action space), and \textit{Efficiency} (optimality without redundancy). This structured analysis is then provided as verbal feedback to contextually guide the agent's error recovery.

\item \textbf{GUI-Critic-R1}~\citep{Wanyan2025guicriticr1} is a specialized, open-source critic model explicitly trained to generate fine-grained verbal feedback for mobile GUI environments. We utilize it as a supervised baseline to compare our zero-shot and test-time scaling approaches against a model that has been directly optimized for trajectory critique and error identification.
\end{itemize}

\myparagraph{Implementation Details.}
We implement our training framework using the \texttt{LlamaFactory}~\citep{zheng2024llamafactory} recipes on 8 NVIDIA H100 GPUs (80GB).
The model is optimized for 1 epoch using a cosine learning rate schedule with a peak learning rate of 5e-6 and a global batch size of 256, taking approximately 4 hours to train.
During test-time deployment, we evaluate the benchmarks using 8 NVIDIA H100 GPUs (80GB). 
We set the history window sizes to $W=4$ and $W=2$ for Qwen3-VL-32B-Thinking and Gemini-3-Flash, respectively, following their official computer-use frameworks~\citep{qwen3-vl, gemini3}. All experimental results are from a single run, due to heavy compute costs for long-horizon GUI tasks. 

\begin{figure*}[t]
    \centering
    \includegraphics[width=\linewidth]{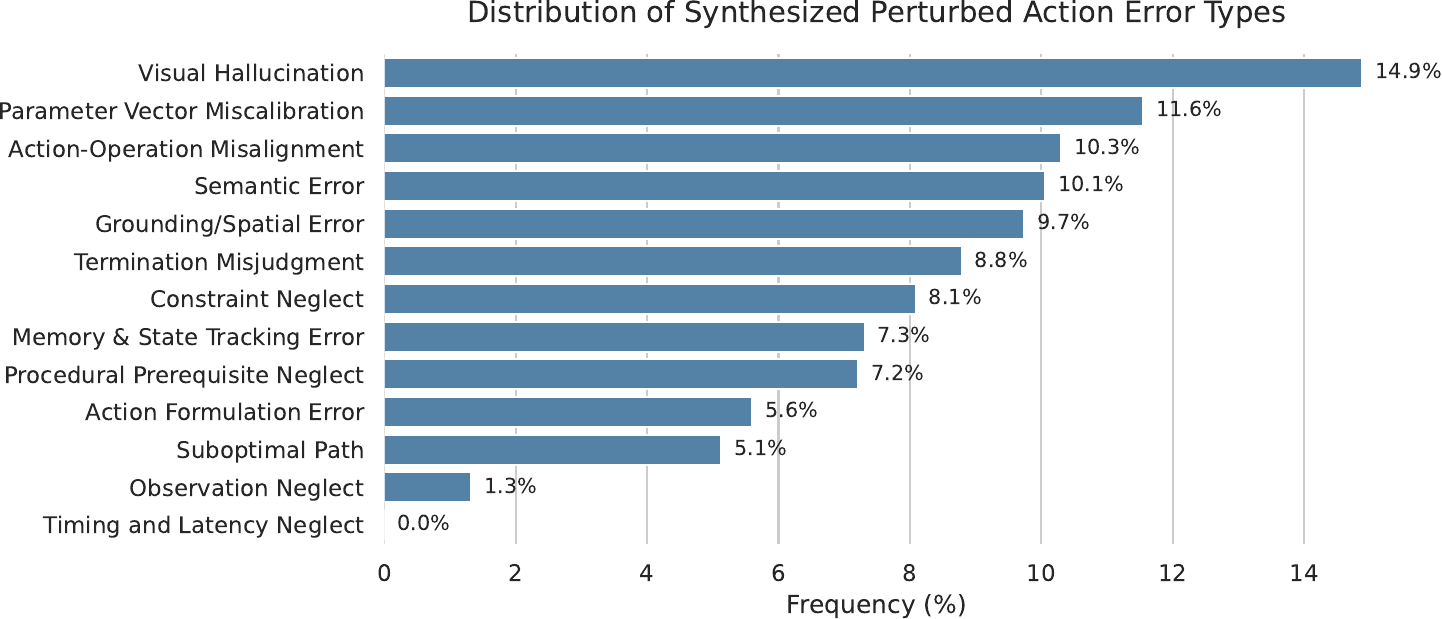}
    \par
    \vspace{-0.05in}
    \caption{\textbf{Distribution of synthesized error types.} We show the distribution of error types within the training dataset for the visually grounded error analysis task.}
    \vspace{-0.05in}
    \label{fig:error_stat}
\end{figure*}

\section{Complete Taxonomy of Synthesized Error Dimensions\label{appendix:Error Dimensions}}
\begin{figure*}[t]
    \centering
    \includegraphics[width=\linewidth]{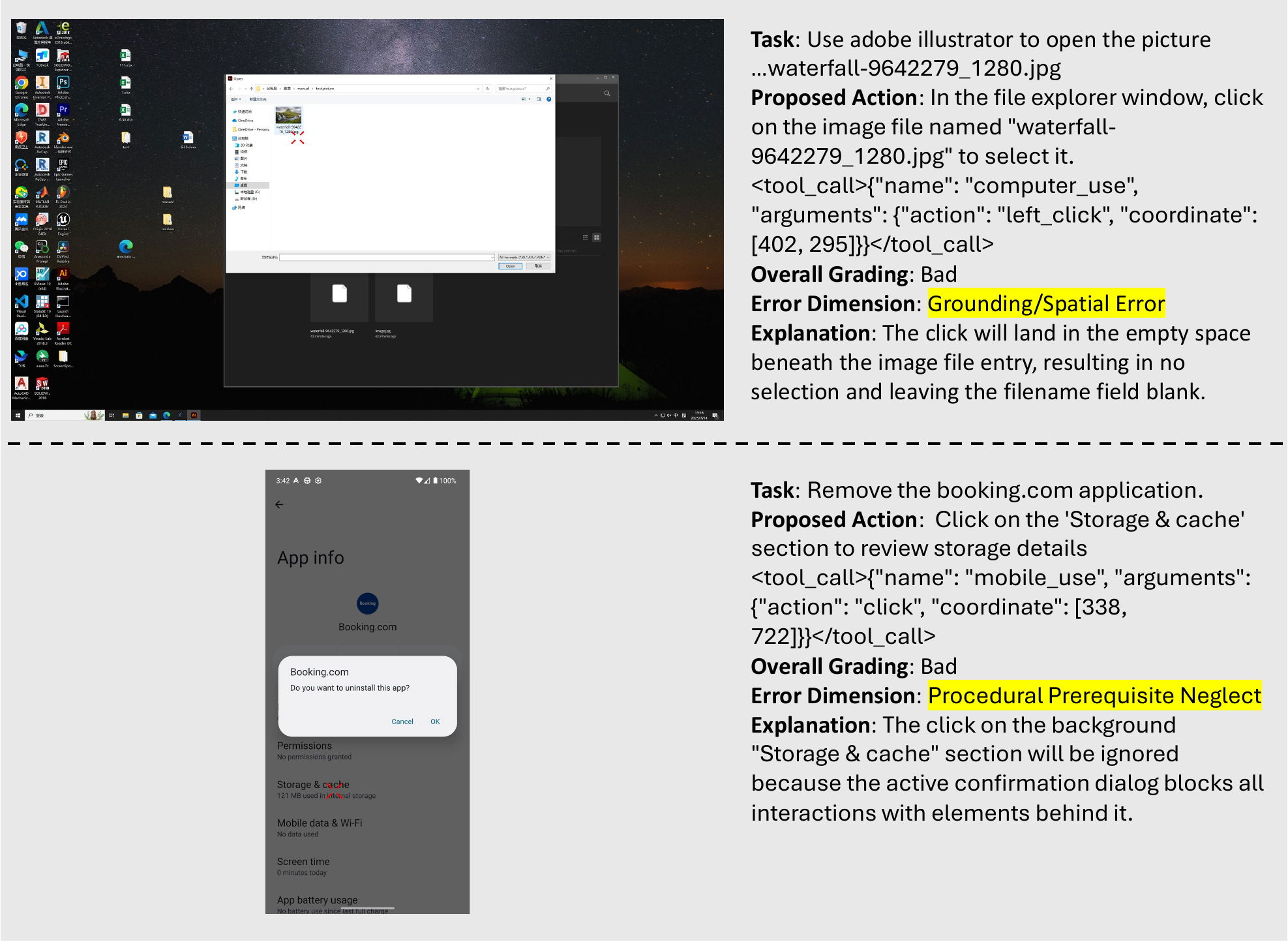}
    \par
    \vspace{-0.075in}
    \caption{Representative examples of common CUA failure modes. (Top) \textit{Grounding/Spatial Error}: The agent forms the correct semantic intent but outputs flawed coordinates, clicking the empty space beneath the file instead of selecting it. (Bottom) \textit{Procedural Prerequisite Neglect}: The agent attempts to interact with a background UI element while a confirmation dialog is active, blocking the click and resulting in a silent failure.}
    \vspace{-0.15in}
    \label{fig:error_example1}
\end{figure*}

\begin{figure*}[t]
    \centering
    \includegraphics[width=\linewidth]{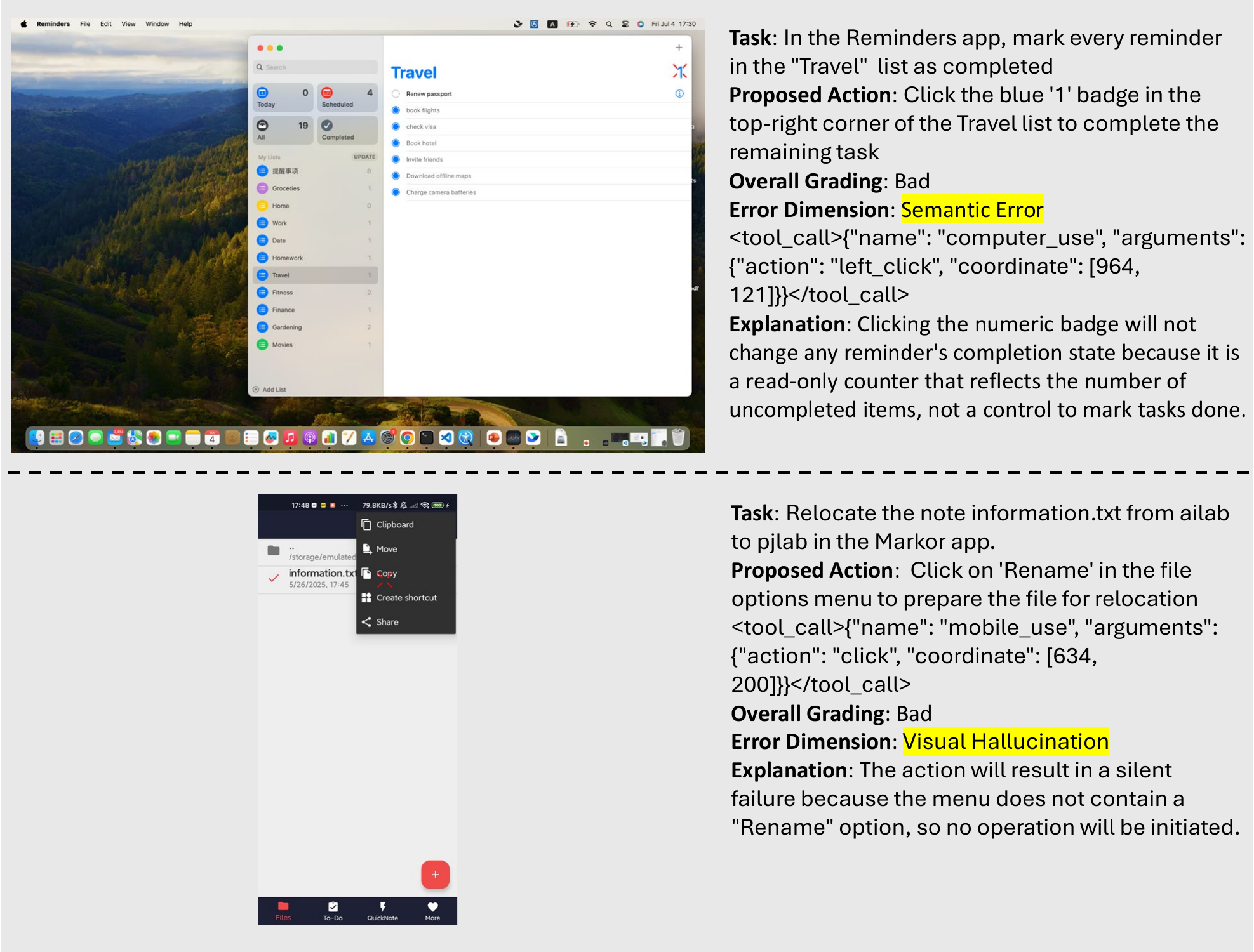}
    \par
    \vspace{-0.075in}
    \caption{Representative examples of common CUA failure modes (continued). (Top) \textit{Semantic Error}: The agent misunderstands UI affordances, attempting to click a read-only numeric counter. (Bottom) \textit{Visual Hallucination}: The agent hallucinates a UI element, attempting to click a "Rename" option that does not exist in the current screenshot.}
    \vspace{-0.15in}
    \label{fig:error_example1_2}
\end{figure*}

\begin{figure*}[t]
    \centering
    \includegraphics[width=\linewidth]{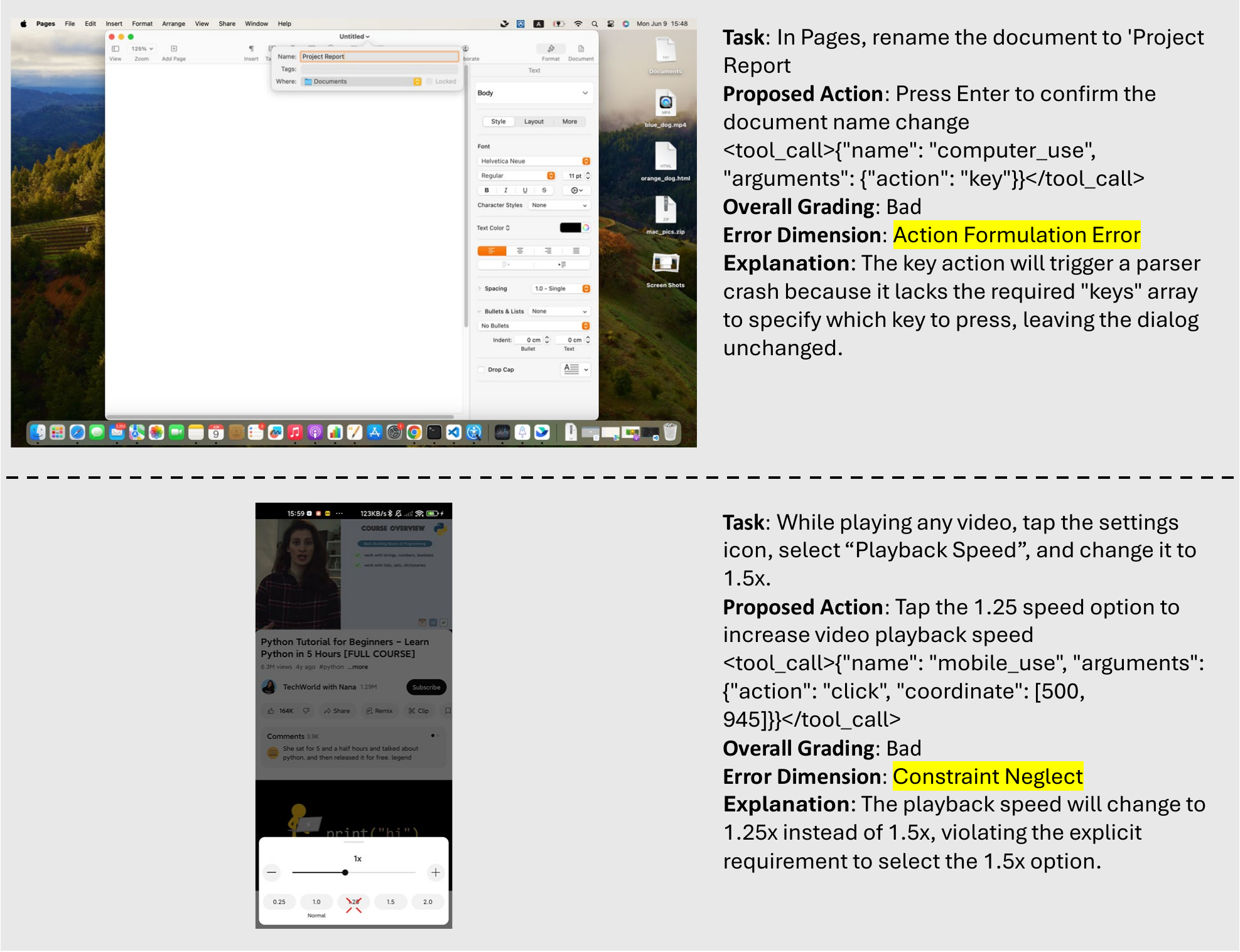}
    \par
    \vspace{-0.075in}
    \caption{Representative examples of common CUA failure modes (continued). (Top) \textit{Action Formulation Error}: The agent intends to press 'Enter' but omits the required key parameter in the JSON payload, crashing the parser. (Bottom) \textit{Constraint Neglect}: The agent successfully navigates the UI but selects a 1.25x playback speed instead of the explicitly requested 1.5x.}
    \vspace{-0.15in}
    \label{fig:error_example2}
\end{figure*}

\begin{figure*}[t]
    \centering
    \includegraphics[width=\linewidth]{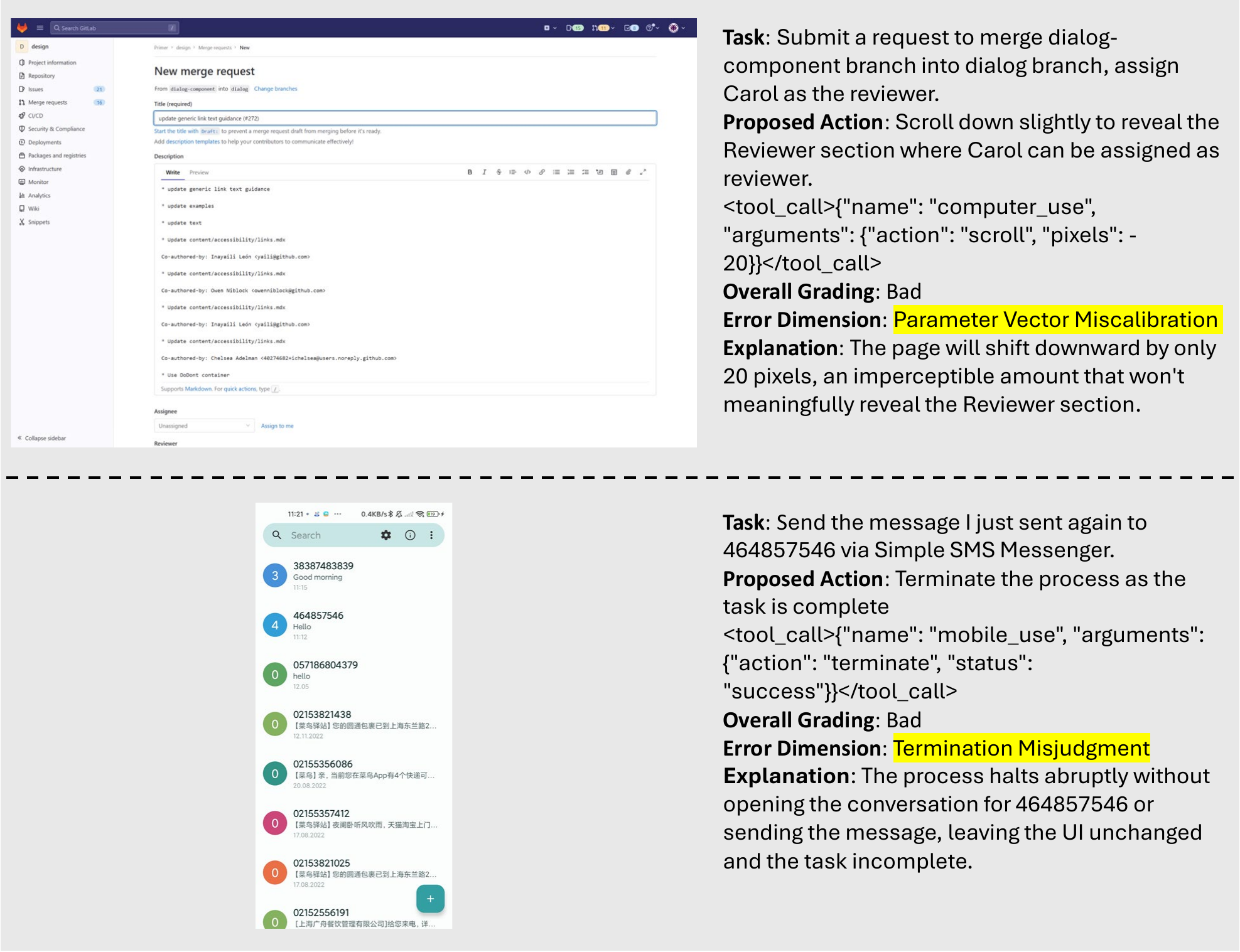}
    \par
    \vspace{-0.075in}
    \caption{Representative examples of common CUA failure modes (continued). (Top) \textit{Parameter Vector Miscalibration}: The agent attempts to scroll to off-screen content but issues a mere 20-pixel magnitude, failing to reveal the target 'Reviewer' section. (Bottom) \textit{Termination Misjudgment}: The agent prematurely halts execution before opening the target conversation and sending the message.}
    \vspace{-0.15in}
    \label{fig:error_example2_2}
\end{figure*}

\begin{figure*}[t]
    \centering
    \includegraphics[width=\linewidth]{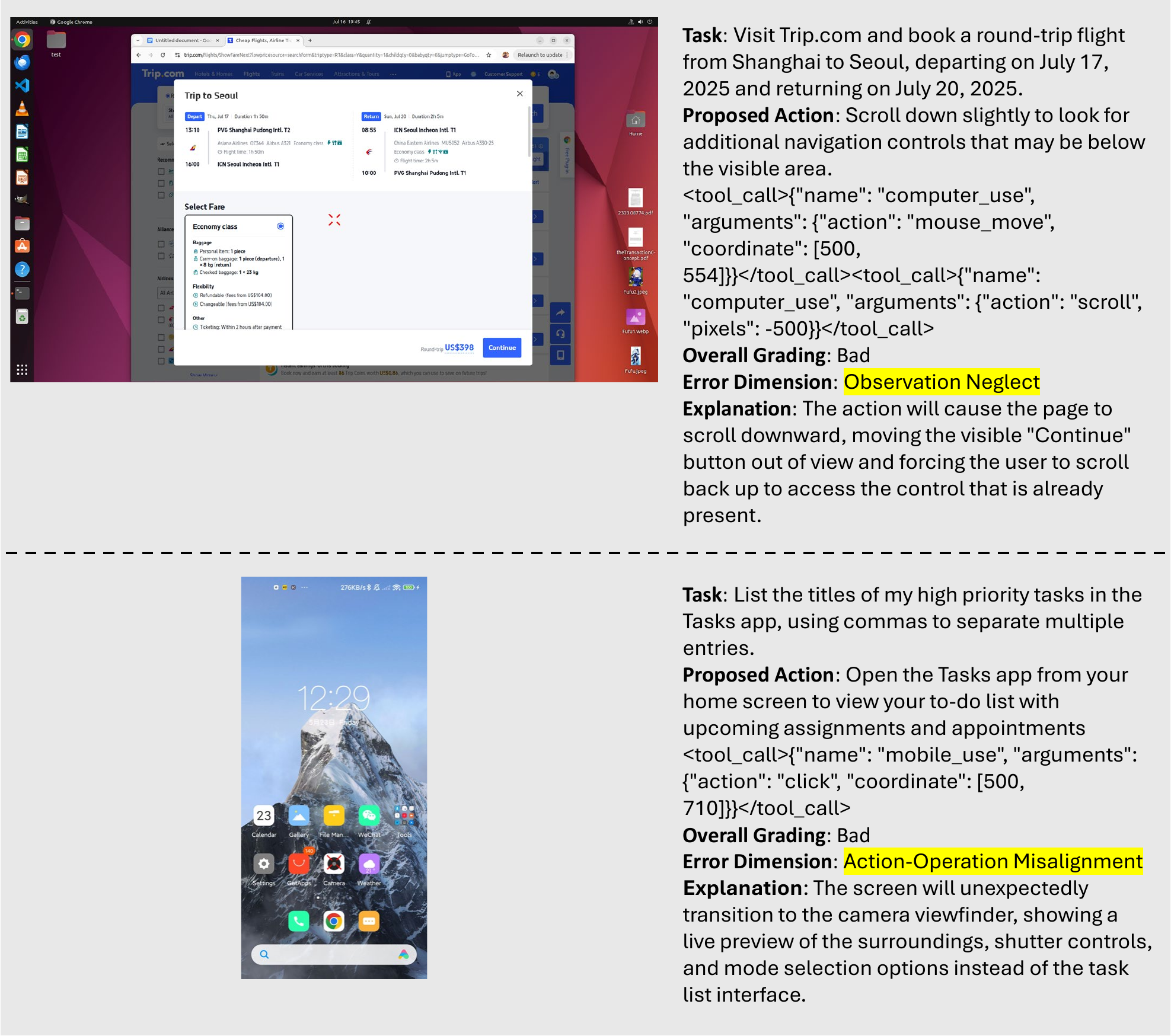}
    \par
    \vspace{-0.075in}
    \caption{Representative examples of common CUA failure modes (continued). (Top) \textit{Observation Neglect}: The agent scrolls downward to search for controls, unnecessarily moving the already visible "Continue" button out of view. (Bottom) \textit{Action-Operation Misalignment}: The agent intends to open the "Tasks" app but outputs coordinates for the camera app, launching the viewfinder instead.}
    \vspace{-0.15in}
    \label{fig:error_example3}
\end{figure*}

\begin{figure*}[t]
    \centering
    \includegraphics[width=\linewidth]{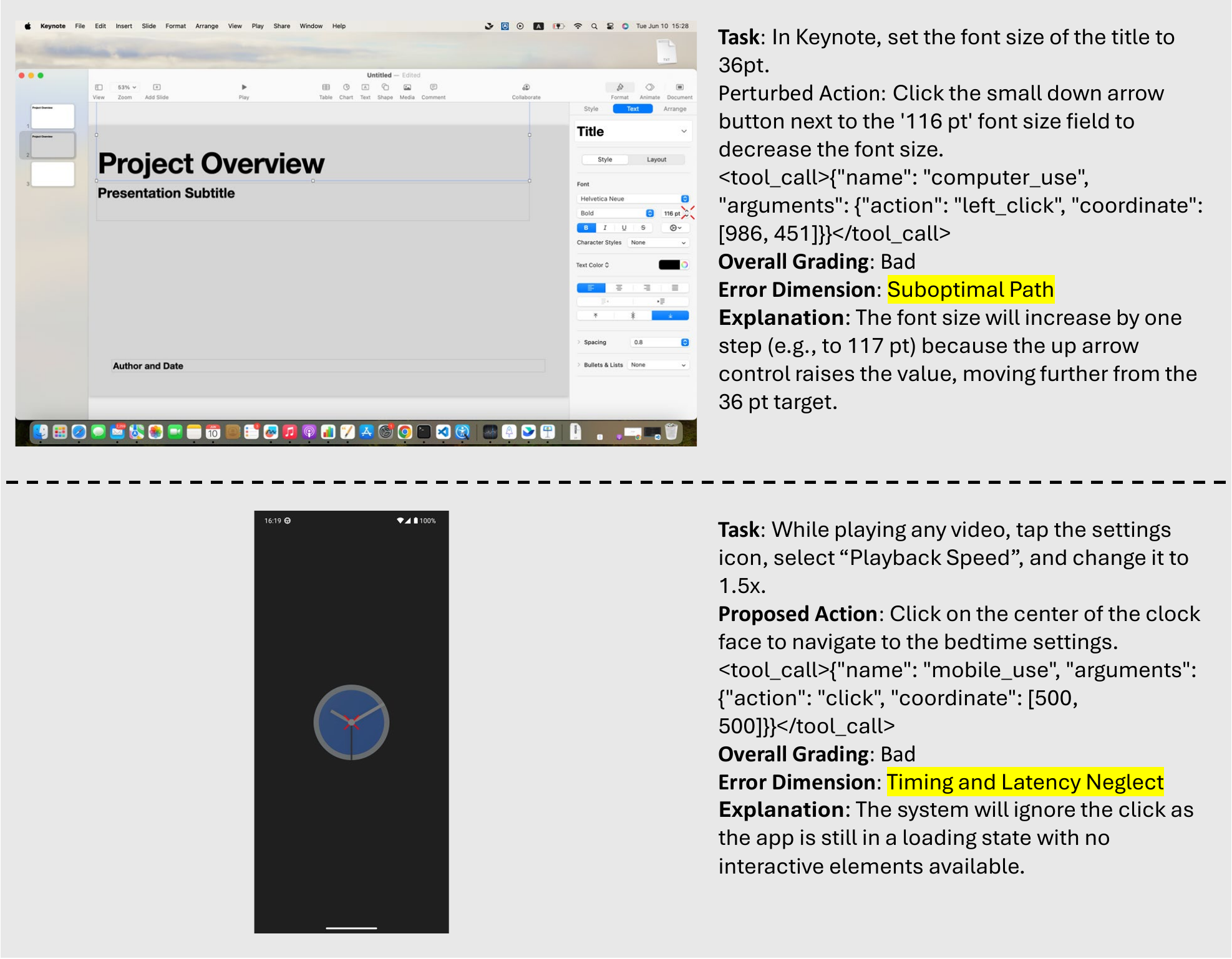}
    \par
    \vspace{-0.075in}
    \caption{Representative examples of common CUA failure modes (continued). (Top) \textit{Suboptimal Path}: The agent attempts to decrease a font size but mistakenly clicks the "up" arrow, moving further from the target goal. (Bottom) \textit{Timing and Latency Neglect}: The agent interacts with the UI during a non-interactive loading state, causing the system to ignore the premature click.}
    \vspace{-0.15in}
    \label{fig:error_example3_2}
\end{figure*}

To ensure our critic learns to detect and recover from a comprehensive set of failure modes, we synthesize plausible errors that mirror the actual mistakes made by standard Computer Use Agents (CUAs). 
Rather than relying on arbitrary perturbations, our taxonomy is empirically based in the primary failure modes observed across recent agent evaluations and benchmarks.
We categorize our 12 error dimensions into two core vulnerabilities of CUAs, directly addressing the limitations outlined in our methodology: Visual and Spatial Grounding and Cognitive Execution.
The distribution of these error dimensions within our constructed training data is illustrated in~\Cref{fig:error_stat}.
Additionally, comprehensive examples for each error dimension are provided across~\Cref{fig:error_example1,fig:error_example1_2,fig:error_example2,fig:error_example2_2,fig:error_example3,fig:error_example3_2}.

\myparagraph{Visual and Spatial Grounding Errors}
These errors occur when the policy's logical reasoning is disconnected from the actual visual reality of the current GUI state, a persistent bottleneck in MLLMs interacting with raw screenshots.
\begin{itemize}[itemsep=2mm, parsep=1pt, leftmargin=*]
    \item \textbf{Grounding/Spatial Error:} The semantic intent is correct, but coordinate precision fails. The agent near-misses the target, landing in adjacent dead space (e.g., a minor pixel offset just outside the bounding box). This is a well-documented failure mode in unconstrained coordinate environments~\citep{Li2025screenspotpro, Wang2025opencua}.
    \item \textbf{Visual Hallucination:} The agent interacts with a non-existent UI element, such as targeting a ghost element from a previous state or guessing a layout coordinate based on parametric priors rather than visual evidence~\citep{Jin2026halluclear}.
    \item \textbf{Observation Neglect:} The agent attempts to search, scroll, or open menus for target elements that are already clearly visible on the current screen~\citep{Jin2026halluclear}.
    \item \textbf{Parameter Vector Miscalibration:} The agent fails physical vector execution~\citep{Rawles2025androidworld}. It either reasons the exact opposite direction of the goal (Polarity Reversal, e.g., the wrong mathematical sign for scrolling) or uses an insufficient scale, resulting in negligible UI movement (Magnitude Insufficiency).
    \item \textbf{Action-Operation Misalignment:} The verbalized intent contradicts the executable JSON string. The reasoned action description is logical, but the actual generated action execution is invalid~\citep{Liu2026scalecua}.
\end{itemize}

\myparagraph{Cognitive and Step-Level Execution Errors}
These errors occur when the agent successfully perceives the screen but fails to reason about the correct immediate action, format, or sequence.
\begin{itemize}[itemsep=2mm, parsep=1pt, leftmargin=*]
    \item \textbf{Termination Misjudgment:} The agent misjudges task completion. It either prematurely outputs a \texttt{terminate} command, fails to explicitly report required data using the \texttt{answer} tool, or hallucinates redundant steps long after the goal has been met~\citep{Wang2025opencua}.
    \item \textbf{Procedural Prerequisite Neglect:} The agent skips a mandatory preceding state change, such as failing to focus a field before typing or failing to dismiss a foreground overlay blocking the target~\citep{Zhou2024webarena, Wang2025opencua}.
    \item \textbf{Semantic Error:} The agent targets perfectly but misinterprets vocabulary, icons, or UI paradigms (e.g., clicking ``Sign Up'' instead of ``Log In'', or clicking a deceptive ad)~\citep{Deng2023mind2web}
    \item \textbf{Constraint Neglect:} The agent ignores a specific attribute or positional constraint explicitly stated in the goal (e.g., selecting the wrong author or wrong position)~\citep{Liu2026scalecua}.
    \item \textbf{Action Formulation Error:} The intent is correct, but the generated JSON crashes the parser due to syntax errors (e.g., missing quotes, trailing commas) or missing required arguments/invalid enums~\citep{Yang2024sweagent}.
    \item \textbf{Suboptimal Path:} The agent selects highly inefficient micro-actions (e.g., repetitive arrow or backspace clicks) instead of standard, faster paradigms (e.g., direct text entry, bulk delete)~\citep{Zhou2024webarena}.
    \item \textbf{Timing and Latency Neglect:} The agent executes an action prematurely, ignoring system busy indicators like loading spinners, unfolding menus, or disabled buttons.
\end{itemize}

\section{Computer Use Agent Action Space\label{appendix:action_space}}
\begin{figure*}[t]
    \centering
    \includegraphics[width=\linewidth]{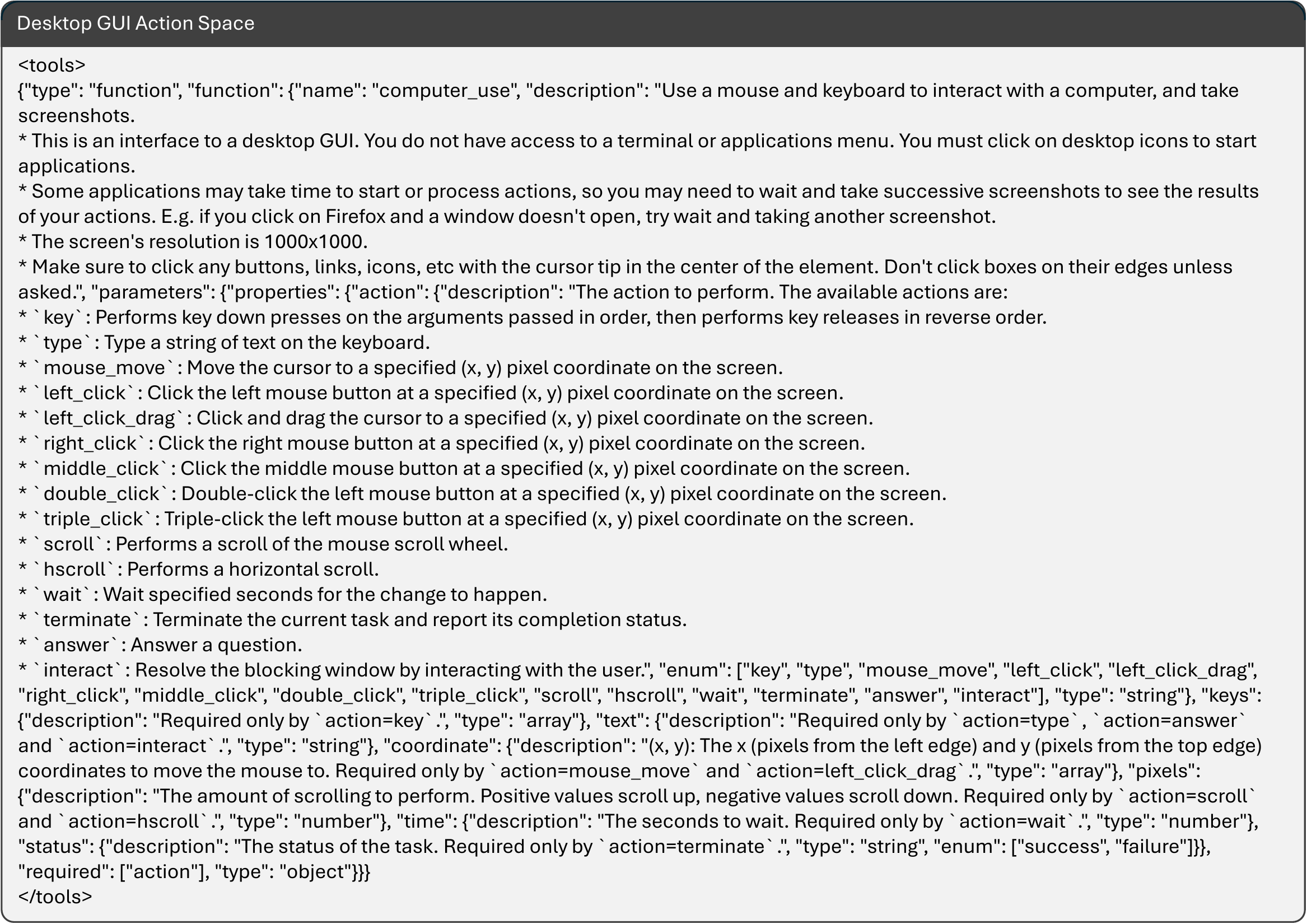}
    \par
    \vspace{-0.075in}
    \caption{Action space definition in desktop (e.g., web, WindowsOS, MacOS) GUI environments.}
    \vspace{-0.15in}
    \label{fig:action_space_desktop}
\end{figure*}

\begin{figure*}[t]
    \centering
    \includegraphics[width=\linewidth]{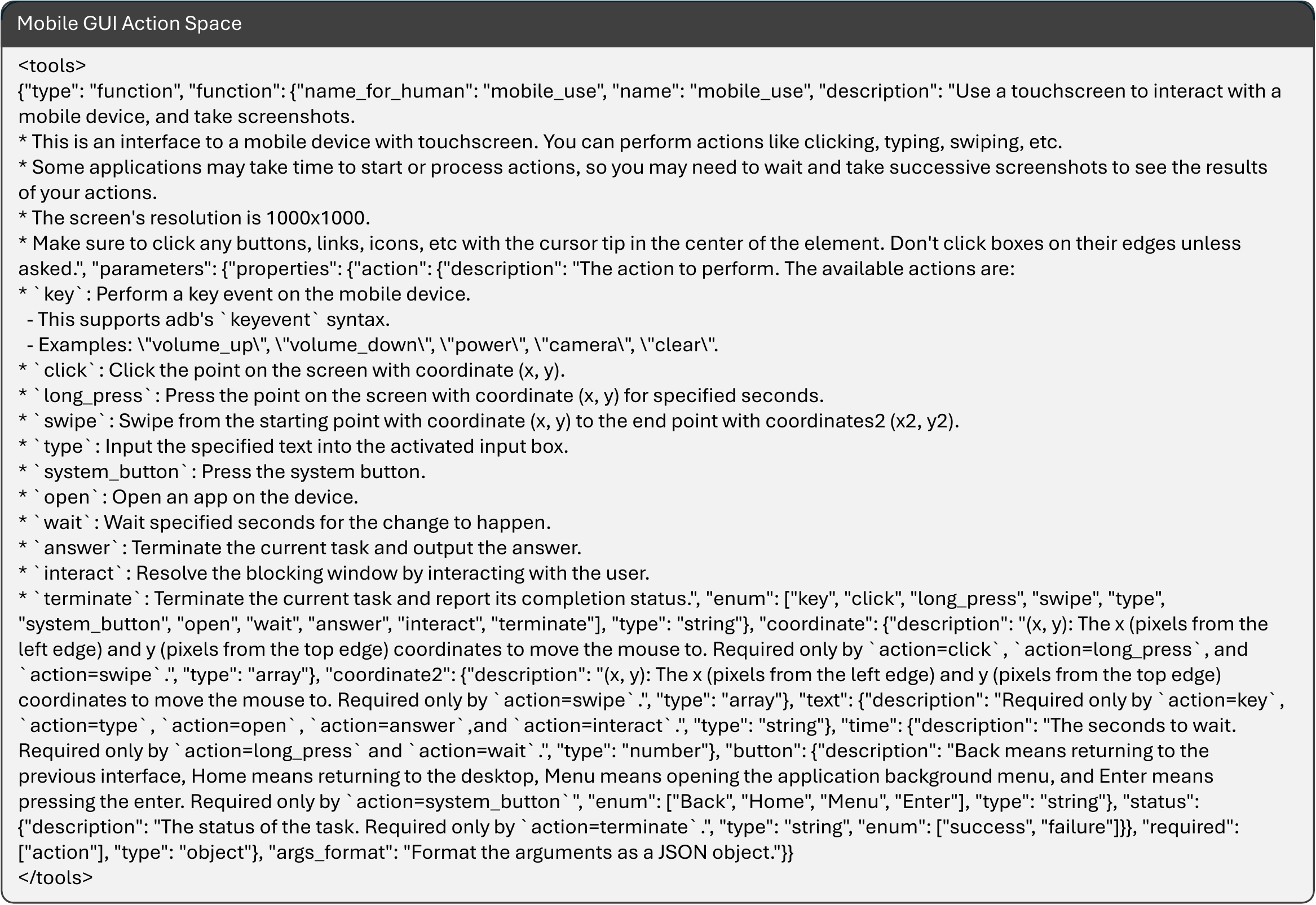}
    \par
    \vspace{-0.075in}
    \caption{Action space definition in mobile GUI environments.}
    \vspace{-0.15in}
    \label{fig:action_space_mobile}
\end{figure*}

To ensure reproducibility, we provide the exact JSON schemas that detail the action spaces, which are adapted from Qwen3-VL~\citep{qwen3-vl}. for both desktop and mobile environments, as shown in~\Cref{fig:action_space_desktop} and~\Cref{fig:action_space_mobile}.
Across all platforms, the visual observation is normalized to a standardized $1000 \times 1000$ coordinate grid.
By establishing this unified continuous pixel space and action grammar, we ensure that both the underlying policies and our test-time critic can generalize their interactions and visually grounded evaluations across diverse GUI environments, independent of the host operating system.

\section{Human Validation of Automated Data Filtering}\label{appendix:human_validation}
In~\Cref{sec:sub:data construction}, we used the automated data filtering pipeline with the multimodal judge (Qwen3-VL-32B) to ensure the quality of the synthesized perturbed actions.
To validate the reliability of our automated double-blind filtering pipeline, we conducted a human evaluation study to ensure the multimodal judge (Qwen3-VL-32B) correctly identifies defects without arbitrary rejections. 
We randomly sampled 50 training instances from our data source.
For each instance, three authors were provided with the visual observation and user instruction, and tasked with identifying the correct expert action between the ground truth and the synthesized perturbed action (with labels blinded).
Using majority vote aggregation, the human consensus achieved a 72.3\% agreement rate with the raw pipeline labels and a 76.6\% agreement rate with the decision made by our automated judge.
This strong correlation demonstrates that our automated filtering aligns closely with human judgment, confirming it as a reliable, scalable mechanism for removing invalid synthesized errors.

\section{Additional Experiments \label{appendix:Additional Experiments}}

\subsection{Lack of Visual Grounding in Previous Critics}\label{appendix:sub:Lack of Visual Grounding in Previous Critics}
\begin{figure*}[t]
    \centering
    \includegraphics[width=\linewidth]{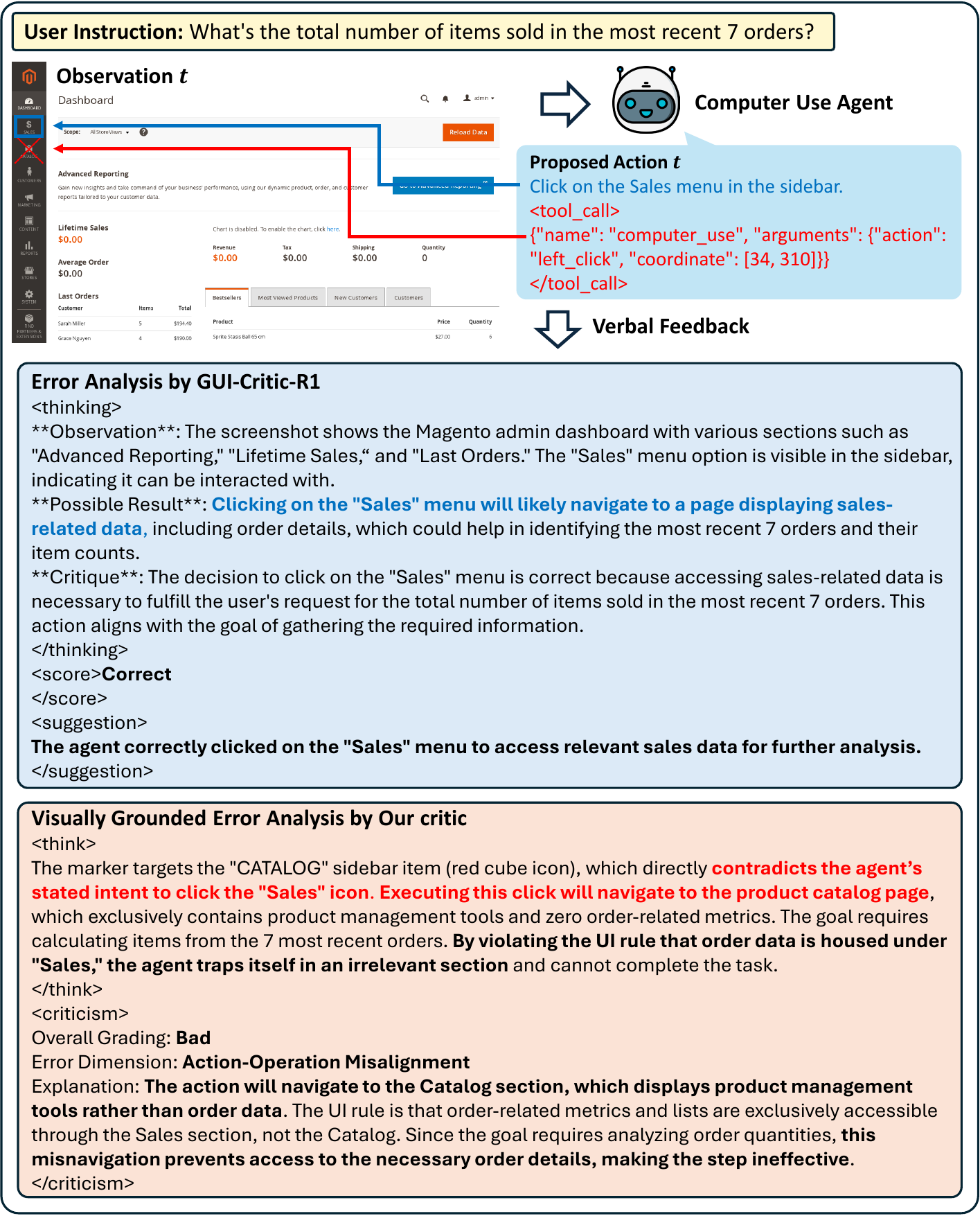}
    \par
    \vspace{-0.075in}
    \caption{Illustration of the visual grounding limitation in existing critics. When the policy generates an action with a logically correct textual intent ("Click on the Sales menu") but outputs misaligned spatial coordinates that actually target the "Catalog" icon, the baseline critic erroneously approves the action by over-relying on the text. In contrast, our proposed critic evaluates the precise physical execution using a visual marker, successfully detecting the spatial mismatch and intercepting the Action-Operation Misalignment before it executes.}
    \vspace{-0.15in}
    \label{fig:visual_grounding}
\end{figure*}

In~\Cref{fig:visual_grounding}, we provide an example where ungrounded GUI-Critic-R1~\citep{Wanyan2025guicriticr1} failed to capture the spatial error while \ours-critic captures the error.
In this scenario, the policy proposes an action with a logically correct textual intent (``Click on the Sales menu'') to fulfill the user's instruction, but outputs incorrect spatial coordinates that actually target the adjacent ``Catalog'' icon.
As GUI-Critic-R1 lacks explicit spatial verification, it over-relies on the agent's verbalized intent.
It assumes the text aligns with the spatial execution, leading it to hallucinate a successful outcome and erroneously approve the flawed action.
In contrast, \ours-critic uses a visual marker on the exact execution coordinates, which helps the critic to verify the actual UI element being targeted.
This allows it to immediately detect the contradiction between the stated intent and the visual reality, correctly detecting an ``Action-Operation Misalignment'' error and providing verbal feedback to the policy to prevent a costly misnavigation.

\subsection{Lack of History State Tracking in Previous Critics}\label{appendix:sub:Lack of History State Tracking in Previous Critics}
\begin{figure*}[t]
    \centering
    \includegraphics[width=\linewidth]{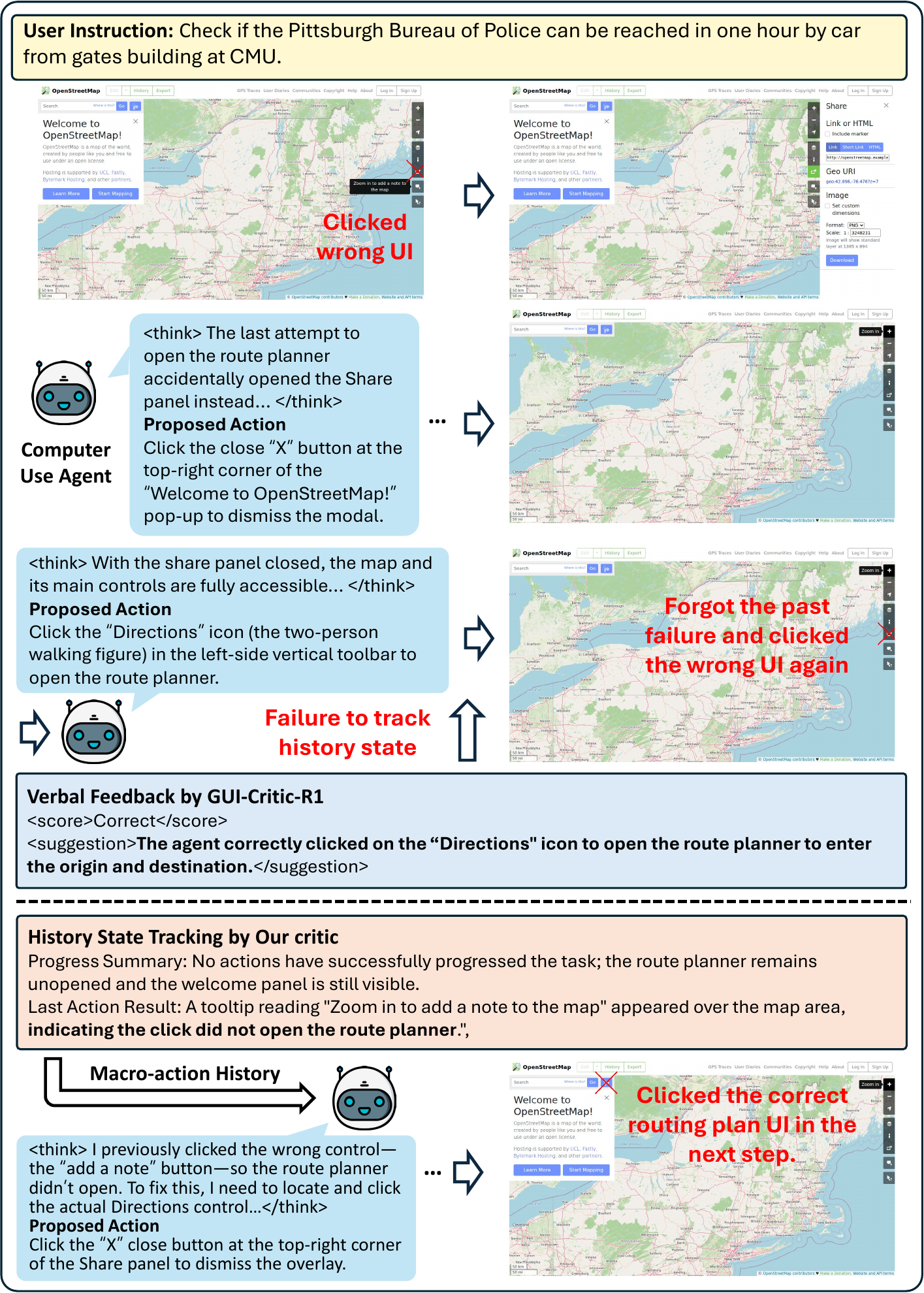}
    \par
    \vspace{-0.075in}
    \caption{Illustration of the history state tracking limitation in existing critics. When the policy generates an action that clicked the wrong UI, the baseline critic fails to log past failures, leaving the policy without the feedback needed to avoid making the same mistake. In contrast, our proposed critic maintains a macro-action history, tracking the history state and enabling the policy to recognize prior failures to proceed in long-horizon GUI task.}
    \vspace{-0.15in}
    \label{fig:history_state_tracking}
\end{figure*}

In~\Cref{fig:history_state_tracking}, we illustrate a failure case where the baseline critic, GUI-Critic-R1~\citep{Wanyan2025guicriticr1}, fails to account for task history, whereas \ours-critic successfully leverages its macro-action history to rectify the agent's behavior.
In this scenario, the agent intends to open a direction planner but performs an incorrect click that fails to trigger the navigation interface. Because the baseline critic lacks a mechanism to track history states, it erroneously labels the action as ``Correct'' and fails to provide any corrective guidance.
This lack of state awareness leaves the policy forget its previous mistake, leading it to repeat the same ineffective click.
In contrast, our critic maintains a macro-action history that logs history states across multiple interactions by the policy.
Providing this marco-action history, our critic allows the policy to recogize the past failure, allowing the policy to make more history-aware decisions to avoid making the same mistake.
This demonstrates that history tracking is essential to guide the policy in long-horizon GUI tasks.

\begin{table}[t]
    \centering
    \small
        \renewcommand{\arraystretch}{1.3}
        \setlength{\tabcolsep}{13pt}
        \begin{tabular}{c c c}
             \toprule
             {\textbf{VisAnalysis}} & {{\textbf{HisTrack}}} & {{\textbf{Overall}}}\\
             \midrule
             Separate \ours & Separate \ours & {\small 23.4} \\
             Separate \ours & Qwen3-VL-8B & {\small 24.7} \\
             Separate \ours & Qwen3-VL-32B & {\small \textbf{25.3}} \\
             \midrule
             \multicolumn{2}{c}{Mixed training \ours (Ours)} & {\small \textbf{25.3}} \\
             \bottomrule
        \end{tabular}
    \vspace{-0.10in}
    \caption{\textbf{Ablation on critic training.} We compare the overall success rate of a unified 8B critic trained on a mixed dataset (Mixed training \ours) against separately trained 8B critics (Separate \ours) and zero-shot baselines that use Qwen3-VL models for historical progress grounding across WebArenaLitev2. VisAnalysis: Visually grounded error analysis, HisTrack: Historical state tracking.
    Both Qwen3-VL-8B and Qwen3-VL-32B are thinking models.
    }
    \label{tab:summary_ablation}
    \vspace{-0.15in}
\end{table}

\begin{figure*}[t]
    \centering
    \includegraphics[width=\linewidth]{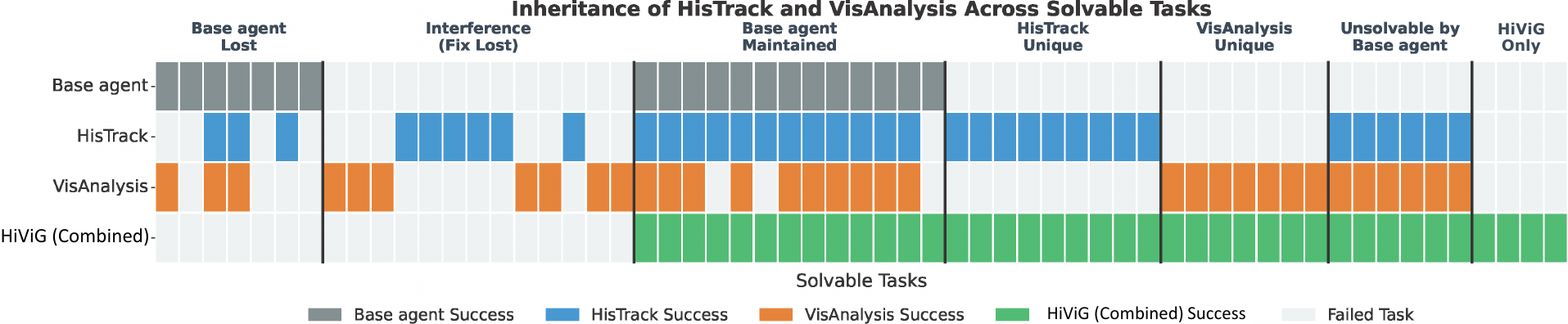}
    \par
    \vspace{-0.05in}
    \caption{\textbf{Task-level success overlap of \ours components.} We visualize the success patterns of the base agent, HisTrack (history state tracking) alone, VisAnalysis (visually grounded error analysis) alone, and the combined \ours framework across all tasks in WebArenaLitev2 solvable by at least one configuration. We use Qwen3-VL-32B-Thinking as the CUA. The presence of distinct tasks uniquely solvable by HisTrack (9 tasks) or VisAnalysis (7 tasks) demonstrates that the two components resolve orthogonal failure modes. The combined framework successfully inherits these independent capabilities—alongside 6 tasks fixed by both—retaining the vast majority of individual fixes. Furthermore, the combined approach exhibits emergent synergy by uniquely solving 4 tasks that all other configurations failed, achieving the highest overall success rate.
    }
    \vspace{-0.05in}
    \label{fig:complementary_component}
\end{figure*}

\subsection{Impact of Mixed Dataset Training and Training Efficiency}\label{appendix:sub:impact of mixed dataset}
We ablate the training data composition by comparing a single critic trained on the mixed dataset against maintaining separate models for each capability, isolating the contribution of each component.
As shown in~\Cref{tab:summary_ablation}, training two separate 8B critics (one exclusively for visually grounded error analysis dataset and another for history state tracking dataset) yields a suboptimal overall success rate of 23.4\% on WebArenaLitev2.
In contrast, jointly training a single model on the mixed dataset increases performance to 25.3\%.
This improvement suggests that visually grounded error analysis and history state tracking are synergic tasks, as they both require pixel-level understanding of GUI screenshots and state-transitions. 
Thus, co-training might provide complementary supervisory signals that enhance the model's overall visual reasoning.
Furthermore, when substituting our critic with the zero-shot history state tracking capabilities of the base Qwen3-VL-8B-Thinking and Qwen3-VL-32B-Thinking models, where the latter is the annotator model, we observe that the 32B model achieves 25.3\% overall success rate.
Our mixed-trained 8B model matches this performance, confirming that our mixed SFT approach successfully distills the history state tracking of the 32B annotator into an efficient, unified 8B critic, minimizing inference overhead without sacrificing accuracy.

Finally, we analyze the impact of training duration.
We compare \ours-critic trained for a single epoch, which is our default configuration, against a model trained for two epochs with our mixed dataset.
Performance remains highly competitive between the two.
The default setting achieves 25.3\% and 51.5\% on WebArenaLitev2 and AndroidLab, respectively, while the two-epoch model reaches 24.7\% and 52.2\%.
This consistency across training durations demonstrates that our framework achieves robust performance with minimal training overhead.

\subsection{Synergistic Effects of \ours's Two Core Components}\label{appendix:sub:synergistic effects of two components}
To further understand how visually grounded error analysis and history state tracking components combine in \ours framework, we analyze the task-level success of the base agent, visually grounded error analysis, history state tracking, and \ours on WebArenaLitev2 when using Qwen3-VL-32B-Thinking as an agent.
As illustrated in~\Cref{fig:complementary_component}, each component independently solves a unique subset of tasks that the base agent cannot, demonstrating that they address orthogonal failure modes.
These independent capabilities are largely inherited by the combined \ours framework, which successfully retains the unique fixes from both individual components.
This synergistic effect is also evident in the Gemini-3-Flash experiments on WebArenaLitev2.
While visually grounded error analysis alone yields a comparably smaller performance increase (from 30.5\% to 35.1\%) compared to the gains from history state tracking alone (from 30.5\% to 42.9\%), integrating both components achieves the highest overall success rate of 45.5\%.
This shows that even when history state tracking acts as the primary driver of performance gains in long-horizon tasks, visually grounded error analysis reliably resolves distinct, complementary errors, making their combination essential for maximizing the agent's overall capabilities. 

\subsection{Impact of State-transitions}\label{appendix:sub:impact of state-transition}
In our data construction pipeline, multimodal rationale grounds the action's consequence in GUI environments by providing the annotator with the ground-truth verbalized state-transition ($v_t$) derived from the actual execution tuple $(o_{t}, a_{t}, o_{t+1})$.
To validate the impact of grounding multimodal rationale on the state-transition, we ablate this component by denying the annotator access to $v_t$, instead letting the annotator predict the state-transition based on its internal parametric knowledge like GUI-Critic-R1~\citep{Wanyan2025guicriticr1}.
Training our critic on this ablation dataset causes a performance drop: the overall success rate falls from 25.3\% to 20.8\% on WebArenaLitev2.
This degradation highlights the difficulty of acting as a reliable world model in complex GUI environments.
Without grounding in actual visual outcomes, the annotator frequently hallucinates incorrect state-transitions, distilling these inaccuracies into the trained critic.
This demonstrates that extracting and learning from verified, ground-truth state-transitions is important to bypass the hallucination issues that can limit prior baseline critics.

\begin{table}[t]
    \centering
    \begin{minipage}{0.48\textwidth}
    \centering
    \small
        \renewcommand{\arraystretch}{1.3}
        \setlength{\tabcolsep}{5.pt}
        \begin{tabular}{c c c c}
             \toprule
             {\textbf{Intent Masking Ratio}} & WALv2 & ALab & Avg. \\
             \midrule
             0\% & {\small 25.3} & {\small 47.1} & {\small 36.2} \\
             30\% & {\small 25.3} & {\small \textbf{51.5}} & {\small \textbf{38.4}} \\
             50\% & {\small \textbf{26.0}} & {\small 49.3} & {\small 37.7} \\
             \bottomrule
        \end{tabular}
    \vspace{-0.10in}
    \caption{\textbf{Ablation on intent masking.} We evaluate the impact of the intent masking ratio for visual grounding. While a 50\% mask yields slight improvements on WebArenaLitev2, the 30\% ratio provides the most robust and balanced performance across platforms. WALv2: WebArenaLitev2, ALab: AndroidLab. Avg.: average overall success rate.
    }
    \label{tab:intent_masking_ablation}
    \end{minipage}
    \vspace{-0.15in}
\end{table}

\subsection{Impact of Intent Masking}\label{appendix:sub:impact of intent masking}
To encourage visual grounding, we incorporate intent masking during data construction.
We evaluate the sensitivity of our framework to this hyperparameter by comparing our default 30\% masking ratio against 0\% (no masking) and 50\% masking.
As shown in~\Cref{tab:intent_masking_ablation}, while increasing the masking ratio to 50\% yields marginal gains on WebArenaLitev2 (26.0\%), it leads to a performance drop on AndroidLab (49.3\%).
Conversely, the absence of intent masking (0\%) results in lower performance across both benchmarks compared to our default setting.
These results demonstrate that while intent masking is essential for enhancing the critic's visual grounding capabilities, our 30\% masking ratio provides the most balanced and competitive performance across diverse GUI environments.
The performance decline at higher masking ratios (50\%) potentially be due to a distributional shift between training and test-time inputs.
While we employ intent masking during training to force visual grounding, the critic receives unmasked intent actions at test time.
Excessive masking during training increases the discrepancy between the training distribution and the test-time objective, causing the critic to struggle to fully understand the given actions.

\subsection{Evaluation on OS-Critic-Bench}\label{appendix:sub:os_critic_bench}
We evaluate \ours-critic on OS-Critic-Bench~\citep{Wu2025osoracle}, which is a benchmark designed to measure how well GUI critics catch erroneous or suboptimal actions proposed by policies.
As shown in~\Cref{tab:os_critic_bench}, while OS-Oracle achieves higher overall performance, it is important to note that OS-Critic-Bench is constructed from the held-out test splits of the data sources used to train OS-Oracle, constituting an in-domain evaluation.
In contrast, \ours-critic is evaluated on OS-Critic-Bench in a strictly zero-shot, out-of-domain setting.
Despite this distribution shift, \ours-critic demonstrates highly competitive performance, notably outperforming OS-Oracle on the web split (65.6\% vs.\ 62.2\% accuracy).

Furthermore, we analyze the intervention behavior of \ours-critic using OS-Critic-Bench. 
Because the long-horizon GUI benchmarks evaluated in our main experiments lack step-by-step ground-truth actions, OS-Critic-Bench, which features human-annotated validity labels for proposed actions, serves as a good proxy 
for human-aligned validation. 
On this benchmark, our critic achieves a 71.4\% overall F1 score, a 1.9\% false rejection rate, and an 89.7\% negative predictive value. 
This demonstrates that \ours-critic acts as a conservative, high-precision filter. 
The extremely low false rejection rate ensures that the policy is rarely derailed from a correct trajectory, while the high negative predictive value guarantees that when our critic does intervene, its corrections are highly reliable. 
This conservative behavior reflects our strict intervention threshold in zero-shot settings, prioritizing non-disruptive support to prevent the critic from injecting erroneous feedback into the policy's execution.

\subsection{Robustness Analysis}\label{appendix:sub:Robustness Analysis}
We also conducted multi-run experiments ($n=3$) to verify the stability of our approach. 
As shown in~\Cref{tab:robustness_analysis}, HiViG consistently maintains state-of-the-art performance with low standard deviations, confirming that our performance gains are robust to execution randomness and environmental noise.

\begin{table}[h]
    \centering
    \small
        \begin{minipage}{0.48\textwidth}
        \renewcommand{\arraystretch}{1.3}
        \setlength{\tabcolsep}{7pt}
        \begin{tabular}{l c c c}
             \toprule
             {\textbf{Method}} & {\textbf{WALv2}} & {\textbf{ALab}} & {\textbf{WAA}}\\
             \midrule
             Base agent & {\small 12.6 (0.8)} & {\small 44.4 (1.2)} & {\small 35.1 (0.6)} \\
             OpenCUA & {\small 15.8 (1.5)} & {\small 46.1 (0.8)} & {\small 35.1 (0.1)} \\
             SE-WSM & {\small 16.5 (0.8)} & {\small 44.9 (1.3)} & {\small 32.0 (0.5)} \\
             CGI (32B) & {\small 16.5 (0.8)} & {\small 44.0 (0.5)} & {\small 34.1 (0.9)} \\
             CGI (8B) & {\small 14.5 (2.0)} & {\small 46.5 (0.7)} & {\small 31.5 (0.8)} \\
             GUI-Critic-R1 & {\small 11.9 (1.6)} & {\small 49.0 (0.4)} & {\small 33.7 (0.9)} \\
             \midrule
             \textbf{\ours (Ours)} & {\small \textbf{25.5 (0.4)}} & {\small \textbf{51.9 (0.4)}} & {\small \textbf{36.7 (0.6)}} \\
             \bottomrule
        \end{tabular}
    \vspace{-0.10in}
    \caption{\textbf{Robustness analysis.} Average success rate (\%) and standard deviation (std) over $n=3$ runs on Qwen3-VL-32B-Thinking. WALv2: WebArenaLitev2, ALab: AndroidLab, WAA: WindowsAgentArena.}
    \label{tab:robustness_analysis}
    \end{minipage}
    \vspace{-0.15in}
\end{table}

\subsection{Failure Case Analysis}\label{appendix:sub:Failure Case Analysis}
While \ours framework significantly improves history awareness and error recovery for policies across diverse domains, our qualitative analysis reveals failure modes stemming from the model's capacity limit in fine-grained visual discrimination.
First, during visually grounded error analysis, the critic occasionally misses nuanced visual discrepancies.
For instance, when an agent is instructed to draw a ``red'' circle in Windows Paint but erroneously targets the ``dark red'' color option, \ours fails to detect this error.
This indicates that while the critic excels at broad visual grounding, it can struggle with highly precise visual feature recognition, such as subtle color variations.
Second, during history state tracking, the critic can sometimes generate inaccurate multi-step achieved goals due to textual bias.
For instance, when the agent attempted to delete an alarm in a mobile environment, the executed action merely hid the alarm's detailed configuration view, leaving the alarm itself present on the screen.
However, mistaking this hidden view for a deleted state, the critic incorrectly recorded a successful deletion in its compressed macro-action history.
Ultimately, both failure modes underscore the need for a more fine-grained understanding of visual features and state-to-state GUI dynamics, suggesting room for further improvements to our test-time intervention framework to continue pushing the performance limits of policies.

\subsection{Detailed performance}\label{appendix:sub:detailed perfro}
Table~\ref{tab:webarenalitev2}, Table~\ref{tab:androidlab}, and Table~\ref{tab:windowsagentarena} show detailed performance of WebArenaliteV2, AndroidLab, and WindowsAgentArena benchmarks, respectively.

\section{Large Language Model~(LLM) Use}
In our research, we employed LLMs for training data construction and writing assistance.
During writing, LLMs were utilized for sentence-level refinement and grammatical polishing. All AI-generated suggestions were carefully reviewed and edited by the authors to maintain the coherency and accuracy.

\begin{table*}[t]
    \small
    \centering
    \renewcommand{\arraystretch}{1.05}
    \setlength{\tabcolsep}{10pt}
    \begin{tabular}{l c c c c c c}
         \toprule
         {\textbf{Method}} & {\textbf{Admin}} & {\textbf{GitLab}} & {\textbf{Shopping}} & {\textbf{Map}} & {\textbf{Reddit}} & {\textbf{Overall}} \\
         \midrule
         \multicolumn{7}{c}{\textit{Computer Use Agent: Qwen3-VL-32B-Thinking}} \\
         \midrule
         Base agent &
         {\small 11.4} & {\small 16.7} & {\small 15.9} & {\small 3.9} & {\small 15.8} & {\small 13.0} \\
         \cmidrule{1-7}
         OpenCUA &
         {\small 14.3} & {\small 16.7} & {\small 20.5} & {\small 3.9} & {\small 15.8} & {\small 14.9} \\
         SE-WSM &
         {\small 11.4} & {\small 16.7} & {\small 15.9} & {\small 3.9} & {\small 5.3} & {\small 11.7} \\
         \cmidrule{1-7}
         CGI (32B) &
         {\small 17.1} & {\small 16.7} & {\small 25.0} & {\small 3.9} & {\small 15.8} & {\small 16.9} \\
         CGI (8B) &
         {\small 14.3} & {\small 13.3} & {\small 13.6} & {\small 3.9} & {\small 15.8} & {\small 12.3} \\
         GUI-Critic-R1 &
         {\small 14.3} & {\small 13.3} & {\small 15.9} & {\small 3.9} & {\small 21.1} & {\small 13.6} \\
         \cmidrule{1-7}
         \cellcolor{gg} \ours (Ours) &
         \cellcolor{gg}{\small 28.6} & \cellcolor{gg}{\small 13.3} & \cellcolor{gg}{\small 31.8} & \cellcolor{gg}{\small 23.1} & \cellcolor{gg}{\small 26.3} & \cellcolor{gg}{\small \textbf{25.3}} \\
         \midrule
         \multicolumn{7}{c}{\textit{Computer Use Agent: Gemini-3-Flash}} \\
         \midrule
          Base agent &
         {\small 25.7} & {\small 56.7} & {\small 25.0} & {\small 19.2} & {\small 26.3} & {\small 30.5} \\
         \cmidrule{1-7}
         OpenCUA &
         {\small 22.9} & {\small 53.3} & {\small 22.7} & {\small 23.1} & {\small 31.6} & {\small 29.9} \\
         SE-WSM &
         {\small 25.7} & {\small 53.3} & {\small 31.8} & {\small 23.1} & {\small 10.5} & {\small 30.5} \\
         \cmidrule{1-7}
         CGI (32B) &
         {\small 37.1} & {\small 33.3} & {\small 31.8} & {\small 19.2} & {\small 21.1} & {\small 29.9} \\
         CGI (8B) &
         {\small 17.1} & {\small 50.0} & {\small 25.0} & {\small 11.5} & {\small 31.6} & {\small 26.6} \\
         GUI-Critic-R1 &
         {\small 8.6} & {\small 36.7} & {\small 27.3} & {\small 15.4} & {\small 21.1} & {\small 22.1} \\
         \cmidrule{1-7}
         \cellcolor{gg} \ours (Ours) &
         \cellcolor{gg}{\small 42.9} & \cellcolor{gg}{\small 70.0} & \cellcolor{gg}{\small 38.6} & \cellcolor{gg}{\small 34.6} & \cellcolor{gg}{\small 42.1} & \cellcolor{gg}{\small \textbf{45.5}} \\
         \bottomrule
    \end{tabular}
    \vspace{-0.05in}
    \caption{Comparison of test-time intervention methods on Qwen3-VL-32B-Thinking and Gemini-3-Flash in WebArenalitev2 benchmark.}
    \label{tab:webarenalitev2}
    \vspace{-0.05in}
\end{table*}
\begin{table*}[t]
    \small
    \centering
    \renewcommand{\arraystretch}{1.05}
    \setlength{\tabcolsep}{2pt}
    \begin{tabular}{l c c c c c c c c c c}
         \toprule
         {\textbf{Method}} & {\textbf{Bluecoins}} & {\textbf{Calendar}} & {\textbf{Cantook}} & {\textbf{Clock}} & {\textbf{Contacts}} & {\textbf{Map}} & {\textbf{Pimusic}} & {\textbf{Setting}} & {\textbf{Zoom}} & {\textbf{Overall}} \\
         \midrule
         \multicolumn{11}{c}{\textit{Computer Use Agent: Qwen3-VL-32B-Thinking}} \\
         \midrule
         Base agent &
         {\small 20.0} & {\small 35.7} & {\small 33.3} & {\small 74.1} & {\small 53.3} & {\small 6.7} & {\small 16.7} & {\small 69.6} & {\small 40.0} & {\small 44.2} \\
         \cmidrule{1-11}
         OpenCUA &
         {\small 33.3} & {\small 28.6} & {\small 41.7} & {\small 70.4} & {\small 53.3} & {\small 6.7} & {\small 25.0} & {\small 73.9} & {\small 60.0} & {\small 47.1} \\
         SE-WSM &
         {\small 20.0} & {\small 42.9} & {\small 41.7} & {\small 70.4} & {\small 53.3} & {\small 13.3} & {\small 8.3} & {\small 69.6} & {\small 60.0} & {\small 45.7} \\
         \cmidrule{1-11}
         CGI (32B) &
         {\small 26.7} & {\small 42.9} & {\small 33.3} & {\small 74.1} & {\small 53.3} & {\small 6.7} & {\small 33.3} & {\small 76.5} & {\small 0.0} & {\small 46.9} \\
         CGI (8B) &
         {\small 26.7} & {\small 42.9} & {\small 25.0} & {\small 77.8} & {\small 53.3} & {\small 0.0} & {\small 16.7} & {\small 69.6} & {\small 20.0} & {\small 44.2} \\
         GUI-Critic-R1 &
         {\small 40.0} & {\small 42.9} & {\small 16.7} & {\small 77.9} & {\small 53.3} & {\small 13.3} & {\small 25.0} & {\small 73.9} & {\small 60.0} & {\small 49.3} \\
         \cmidrule{1-11}
         \cellcolor{gg}\ours (Ours) &
         \cellcolor{gg}{\small 40.0} & \cellcolor{gg}{\small 42.9} & \cellcolor{gg}{\small 50.0} & \cellcolor{gg}{\small 70.4} & \cellcolor{gg}{\small 60.0} & \cellcolor{gg}{\small 13.3} & \cellcolor{gg}{\small 16.7} & \cellcolor{gg}{\small 78.3} & \cellcolor{gg}{\small 60.0} & \cellcolor{gg}{\small \textbf{51.5}} \\
         \midrule
         \multicolumn{11}{c}{\textit{Computer Use Agent: Gemini-3-Flash}} \\
         \midrule
          Base agent &
         {\small 66.7} & {\small 42.9} & {\small 75.0} & {\small 74.1} & {\small 53.3} & {\small 13.3} & {\small 25.0} & {\small 78.3} & {\small 80.0} & {\small 58.0} \\
         \cmidrule{1-11}
         OpenCUA &
         {\small 60.0} & {\small 42.9} & {\small 50.0} & {\small 81.5} & {\small 26.7} & {\small 40.0} & {\small 33.3} & {\small 78.3} & {\small 80.0} & {\small 57.2} \\
         SE-WSM &
         {\small 53.3} & {\small 42.9} & {\small 58.3} & {\small 77.9} & {\small 40.0} & {\small 40.0} & {\small 33.3} & {\small 73.9} & {\small 80.0} & {\small 57.2} \\
         \cmidrule{1-11}
         CGI (32B) &
         {\small 60.0} & {\small 42.9} & {\small 50.0} & {\small 74.1} & {\small 33.3} & {\small 40.0} & {\small 33.3} & {\small 78.3} & {\small 60.0} & {\small 55.8} \\
         CGI (8B) &
         {\small 60.0} & {\small 35.7} & {\small 58.3} & {\small 88.9} & {\small 26.7} & {\small 26.7} & {\small 33.3} & {\small 73.9} & {\small 40.0} & {\small 55.1} \\
         GUI-Critic-R1 &
         {\small 60.0} & {\small 42.9} & {\small 50.0} & {\small 66.7} & {\small 46.7} & {\small 26.7} & {\small 25.0} & {\small 73.9} & {\small 80.0} & {\small 53.6} \\
         \cmidrule{1-11}
         \cellcolor{gg}\ours (Ours) &
         \cellcolor{gg}{\small 73.3} & \cellcolor{gg}{\small 42.9} & \cellcolor{gg}{\small 50.0} & \cellcolor{gg}{\small 88.9} & \cellcolor{gg}{\small 40.0} & \cellcolor{gg}{\small 40.0} & \cellcolor{gg}{\small 33.3} & \cellcolor{gg}{\small 78.3} & \cellcolor{gg}{\small 80.0} & \cellcolor{gg}{\small \textbf{61.6}} \\
         \bottomrule
    \end{tabular}
    \vspace{-0.05in}
    \caption{Comparison of test-time intervention methods on Qwen3-VL-32B-Thinking and Gemini-3-Flash in AndroidLab benchmark.}
    \label{tab:androidlab}
    \vspace{-0.05in}
\end{table*}
\begin{table*}[t]
    \small
    \centering
    \renewcommand{\arraystretch}{1.05}
    \setlength{\tabcolsep}{4pt}
    \begin{tabular}{l c c c c c c c}
         \toprule
         {\textbf{Method}} & {\textbf{Office}} & {\textbf{Web Browsing}} & {\textbf{Windows System}} & {\textbf{Code}} & {\textbf{Media}} & {\textbf{Windows Utilities}} & {\textbf{Overall}} \\
         \midrule
         \multicolumn{8}{c}{\textit{Computer Use Agent: Qwen3-VL-32B-Thinking}} \\
         \midrule
         Base agent &
         {\small 4.7} & {\small 57.1} & {\small 50.0} & {\small 58.3} & {\small 27.7} & {\small 50.0} & {\small 35.7} \\
         \cmidrule{1-8}
         OpenCUA &
         {\small 2.3} & {\small 61.9} & {\small 70.8} & {\small 50.0} & {\small 24.5} & {\small 25.0} & {\small 35.2} \\
         SE-WSM &
         {\small 2.3} & {\small 52.4} & {\small 58.3} & {\small 50.0} & {\small 22.9} & {\small 25.0} & {\small 31.6} \\
         \cmidrule{1-8}
         CGI (32B) &
         {\small 4.7} & {\small 47.6} & {\small 58.3} & {\small 50.0} & {\small 37.2} & {\small 25.0} & {\small 33.7} \\
         CGI (8B) &
         {\small 4.7} & {\small 47.6} & {\small 58.3} & {\small 41.7} & {\small 28.1} & {\small 25.0} & {\small 31.0} \\
         GUI-Critic-R1 &
         {\small 4.7} & {\small 61.9} & {\small 45.8} & {\small 54.2} & {\small 28.1} & {\small 41.7} & {\small 34.4} \\
         \cmidrule{1-8}
         \cellcolor{gg}\ours (Ours) &
         \cellcolor{gg}{\small 7.0} & \cellcolor{gg}{\small 57.1} & \cellcolor{gg}{\small 58.3} & \cellcolor{gg}{\small 58.3} & \cellcolor{gg}{\small 28.9} & \cellcolor{gg}{\small 50.0} & \cellcolor{gg}{\small \textbf{38.0}} \\
         \midrule
         \multicolumn{8}{c}{\textit{Computer Use Agent: Gemini-3-Flash}} \\
         \midrule
         Base agent &
         {\small 4.7} & {\small 52.4} & {\small 58.3} & {\small 45.8} & {\small 32.9} & {\small 58.3} & {\small 35.8} \\
         \cmidrule{1-8}
         OpenCUA &
         {\small 4.7} & {\small 52.4} & {\small 58.3} & {\small 45.8} & {\small 32.9} & {\small 58.3} & {\small 35.8} \\
         SE-WSM &
         {\small 4.7} & {\small 42.9} & {\small 62.5} & {\small 45.8} & {\small 28.1} & {\small 66.7} & {\small 35.1} \\
         \cmidrule{1-8}
         CGI (32B) &
         {\small 4.7} & {\small 47.6} & {\small 62.5} & {\small 58.3} & {\small 32.9} & {\small 50.0} & {\small 37.9} \\
         CGI (8B) &
         {\small 4.7} & {\small 57.1} & {\small 62.5} & {\small 54.2} & {\small 37.7} & {\small 41.7} & {\small 37.9} \\
         GUI-Critic-R1 &
         {\small 2.3} & {\small 57.1} & {\small 58.3} & {\small 45.8} & {\small 19.3} & {\small 41.7} & {\small 32.5} \\
         \cmidrule{1-8}
         \cellcolor{gg}\ours (Ours) &
         \cellcolor{gg}{\small 23.3} & \cellcolor{gg}{\small 57.1} & \cellcolor{gg}{\small 62.5} & \cellcolor{gg}{\small 54.2} & \cellcolor{gg}{\small 28.9} & \cellcolor{gg}{\small 66.7} & \cellcolor{gg}{\small \textbf{44.2}} \\
         \bottomrule
    \end{tabular}
    \vspace{-0.05in}
    \caption{Comparison of test-time intervention methods on Qwen3-VL-32B-Thinking and Gemini-3-Flash in WindowsAgentArena benchmark.}
    \label{tab:windowsagentarena}
    \vspace{-0.05in}
\end{table*}
\begin{table*}[t]
    \centering
    \begin{minipage}{\textwidth}
    \centering
    \small
        \renewcommand{\arraystretch}{1.3}
        \setlength{\tabcolsep}{12.pt}
        \begin{tabular}{l c c c c c c c c}
             \toprule
             \multirow{2}{*}{\textbf{Model}} & \multicolumn{2}{c}{\textbf{Mobile}} & \multicolumn{2}{c}{\textbf{Desktop}} & \multicolumn{2}{c}{\textbf{Web}} & \multicolumn{2}{c}{\textbf{Overall}} \\
             \cmidrule(lr){2-3} \cmidrule(lr){4-5} \cmidrule(lr){6-7} \cmidrule(lr){8-9}
             & Acc & F1 & Acc & F1 & Acc & F1 & Acc & F1 \\
             \midrule
             Qwen2.5-VL-7B & {\small 58.5} & {\small 65.8} & {\small 61.2} & {\small 69.4} & {\small 54.7} & {\small 64.2} & {\small 58.3} & {\small 66.2} \\
             Qwen3-VL-8B & {\small 61.6} & {\small 62.3} & {\small 59.9} & {\small 63.9} & {\small 60.1} & {\small 60.4} & {\small 61.0} & {\small 62.3} \\
             UI-TARS-1.5-7B & {\small 56.2} & {\small 66.4} & {\small 58.6} & {\small 69.6} & {\small 60.1} & {\small 70.1} & {\small 57.5} & {\small 67.8} \\
             ScaleCUA-7B & {\small 50.2} & {\small 59.9} & {\small 44.7} & {\small 58.8} & {\small 34.5} & {\small 49.2} & {\small 45.9} & {\small 57.5} \\
             ScaleCUA-32B & {\small 51.6} & {\small 63.3} & {\small 49.3} & {\small 64.8} & {\small 50.7} & {\small 65.4} & {\small 51.0} & {\small 64.1} \\
             SE-WSM-7B & {\small 50.9} & {\small 17.0} & {\small 50.7} & {\small 13.8} & {\small 41.2} & {\small 18.7} & {\small 48.9} & {\small 16.8} \\
             GUI-Critic-R1 & {\small 59.1} & {\small 66.5} & {\small 59.2} & {\small 70.2} & {\small 60.8} & {\small 72.9} & {\small 59.5} & {\small 68.8} \\
             OS-Oracle-7B-SFT & {\small 63.5} & {\small 71.2} & {\small 63.2} & {\small 72.0} & {\small 62.2} & {\small 71.7} & {\small 63.1} & {\small 71.5} \\
             OS-Oracle-7B & {\small \textbf{70.8}} & {\small \textbf{74.3}} & {\small \textbf{65.8}} & {\small \textbf{71.1}} & {\small 62.2} & {\small 70.5} & {\small \textbf{68.0}} & {\small \textbf{72.8}} \\
             \midrule
             \ours-critic (Ours) & {\small 57.0} & {\small 69.7} & {\small 57.9} & {\small 70.1} & {\small \textbf{65.5}} & {\small \textbf{76.1}} & {\small 59.3} & {\small 71.4} \\
             \bottomrule
        \end{tabular}
    \vspace{-0.05in}
    \caption{Comparison of critic performance on OS-Critic-Bench. Acc and F1 denote Accuracy and F1-score, respectively. Baseline performances are from \citet{Wu2025osoracle}.}
    \label{tab:os_critic_bench}
    \end{minipage}
    \vspace{-0.05in}
\end{table*}

\begin{figure*}[t]
    \centering
    \includegraphics[width=\linewidth]{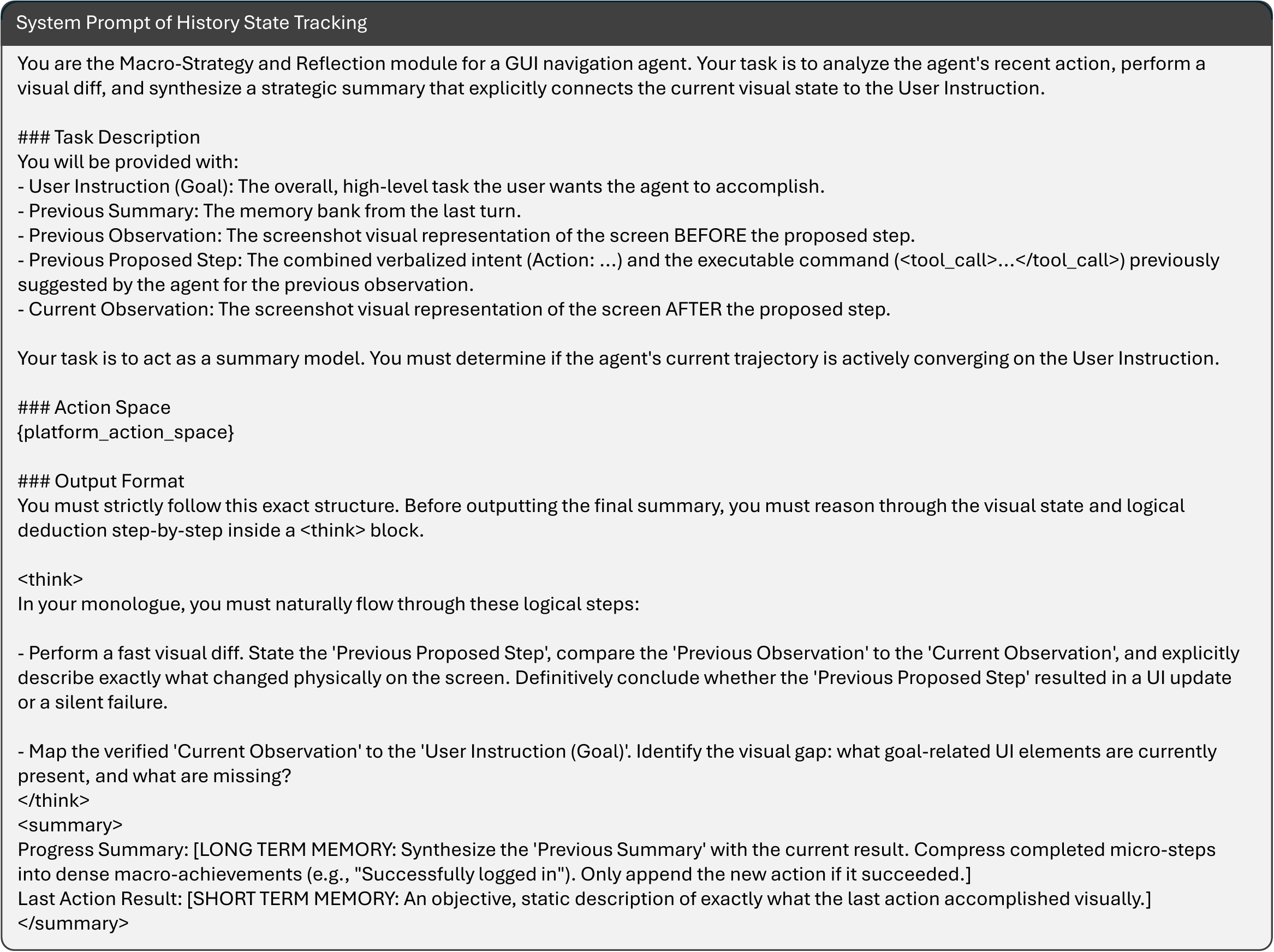}
    \par
    \vspace{-0.075in}
    \caption{Input prompt for \ours for history state tracking at test-time.}
    \vspace{-0.15in}
    \label{fig:system_prompt_history_state_tracking}
\end{figure*}

\begin{figure*}[t]
    \centering
    \includegraphics[width=\linewidth]{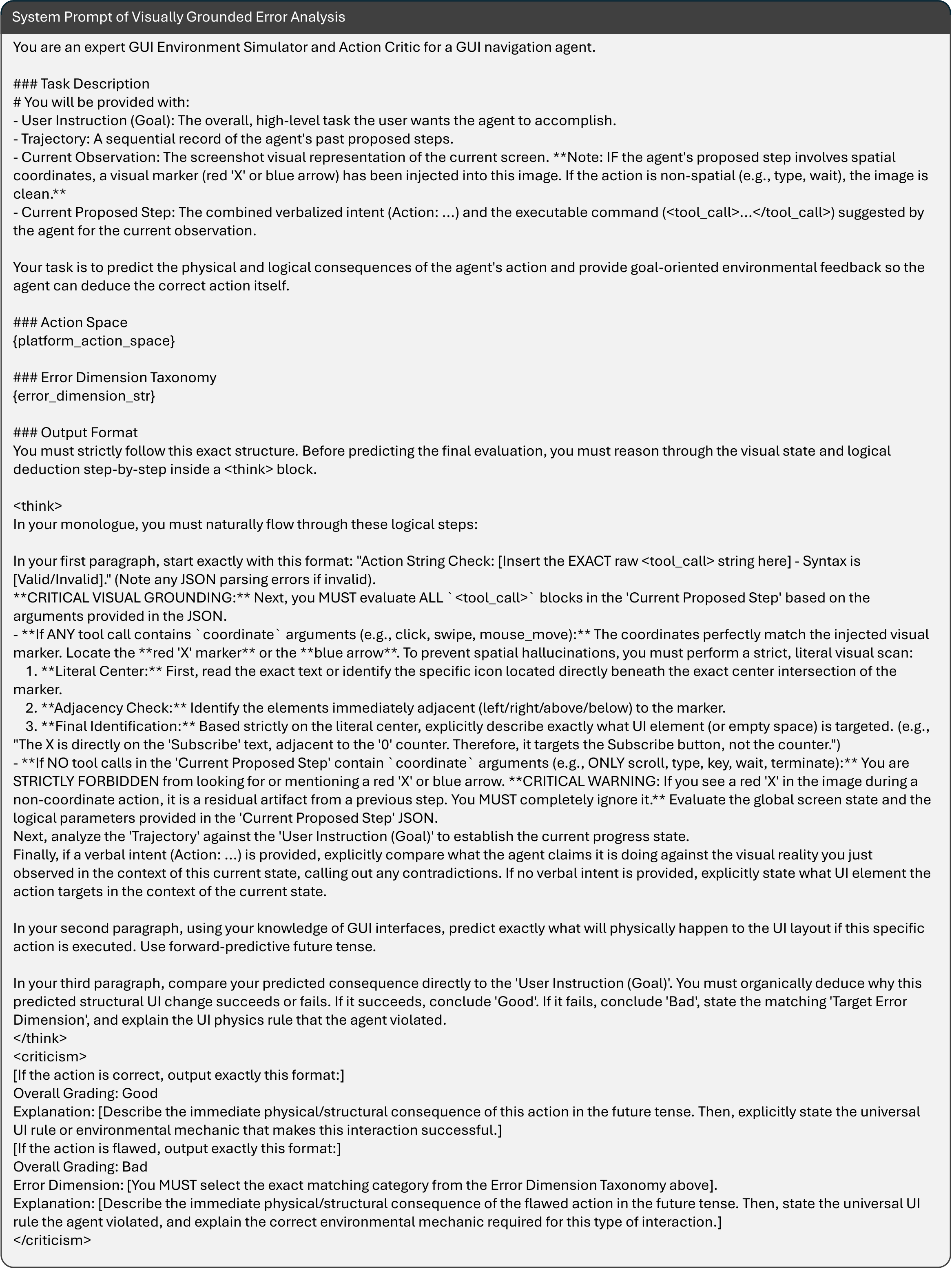}
    \par
    \vspace{-0.075in}
    \caption{Input prompt for \ours for visually-grounded error analysis at test-time.}
    \vspace{-0.15in}
    \label{fig:system_prompt_visually_grounded_error_analysis}
\end{figure*}

\end{document}